%% file: main.tex
\documentclass{article}

\usepackage{preprint}

\usepackage{hyperref}
\usepackage{url}

\usepackage{latexsym}
\usepackage{microtype}
\usepackage{graphicx}
\usepackage{wrapfig}
\usepackage{booktabs}
\usepackage{amsmath}
\usepackage{amssymb}
\usepackage{fvextra}
\usepackage{xcolor}
\usepackage{fontspec}
\usepackage{xeCJK}
\usepackage{enumitem}
\setlist[itemize]{leftmargin=*,topsep=0pt,partopsep=0pt,parsep=0pt,itemsep=0pt}
\setlist[enumerate]{leftmargin=*,topsep=0pt,partopsep=0pt,parsep=0pt,itemsep=0pt}

\setCJKmonofont{FandolSong-Regular.otf}

\input{src/macro}

\title{Do Proactive Agents Need an LLM to Decide \\When to Act?}

\author{
Xiaoze Liu$^{1}$ \quad
Ruowang Zhang$^{1}$ \quad Amir H. Abdi$^{2}$ \quad Michel Galley$^{2}$ \\
{\bfseries Zhikai Chen$^{3}$ \quad Siheng Xiong$^{4}$ \quad Xiaoqian Wang$^{1}$ \quad Jing Gao$^{1}$} \\
$^{1}$Purdue University \quad $^{2}$Microsoft \\
$^{3}$Michigan State University \quad $^{4}$Georgia Institute of Technology
}

\hypersetup{
  hidelinks,
  pdftitle={Do Proactive Agents Need an LLM to Decide When to Act?},
  pdfauthor={Xiaoze Liu, Ruowang Zhang, Amir H. Abdi, Michel Galley, Zhikai Chen, Siheng Xiong, Xiaoqian Wang, Jing Gao},
  pdfsubject={Preprint},
  pdfkeywords={proactive agents, temporal graph learning, graph neural networks}
}

\begin{document}
\maketitle

\begin{abstract}
\input{src/0_abstract}
\end{abstract}

\begin{figure}[!h]
    \centering
    \includegraphics[width=\textwidth]{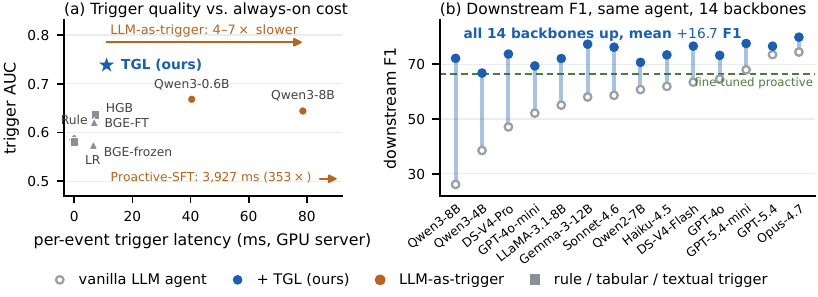}
    \caption{A lightweight graph controller supplies the always-on trigger and context-selection interface. \textbf{(a)} \tgl{} achieves the highest trigger AUC among nine architectures at $11$\,ms per event; the LLM trigger protocols cost $4$--$353\times$ as much (Section~\ref{sec:results:trigger-arch}). \textbf{(b)} The controller improves F1 across all $14$ downstream backbones by a mean of $16.7$ points and exceeds both published fine-tuned proactive baselines on their corresponding frozen backbones.}
    \label{fig:teaser}
\end{figure}

\input{src/1_intro}
\input{src/2_related}

\input{src/3_problem}

\input{src/4_method}
\input{src/5_experiment}

\input{src/10_conclusion}

\input{src/K_ethics}
\input{src/N_reproducibility}
\input{src/L_llm_use}

\input{main.bbl}
\appendix
\input{src/A_fingertip}
\input{src/B_graph_construction}
\input{src/C_downstream_agent}
\input{src/D_additional_experiments}
\input{src/M_rebuttal_analyses}
\input{src/E_implementation_details}
\input{src/F_prompts}
\input{src/G_case_study}
\input{src/H_discussion}

\end{document}

%% file: src/macro.tex
\newcommand{\tgl}{\textsc{TGL}}

%% file: src/0_abstract.tex
Proactive assistants continuously decide when to intervene and what context should support the intervention. Large language model (LLM) pipelines repeatedly interpret activity histories to make these decisions, paying an inference cost even when the assistant remains silent. We show that a small graph model can handle both decisions and improve the language agents it controls. Our key insight is that user activity has a native graph structure: events involve persistent entities whose recurrence connects interactions over time. Triggering and context selection map directly to predictions on event and entity nodes. We introduce a temporal-graph-learning (\tgl{}) controller that learns these predictions jointly and supplies both outputs in one forward pass. The downstream language agent generates suggestions on triggered events using the activity history and scored entities. At $11.13$\,ms per event on a GPU server, \tgl{} achieves the highest AUCs among nine trigger architectures and gives $\sim\!4$--$7\times$ trigger-stage speedups over the two single-forward LLM triggers. A shared \tgl{} model improves F1 across all $14$ downstream backbones by a mean of $16.7$ points. The controller also runs at $13.99$\,ms on a consumer laptop with a ${\sim}220$\,MiB BF16 resident footprint, bringing effective proactive control to on-device deployment.

%% file: src/1_intro.tex
\section{Introduction}
\label{sec:intro}

Proactive assistants bring large language model (LLM) agents into a user's ongoing workflow, offering help as opportunities arise across desktop and mobile activity \citep{lu2025proactiveagent,yang2025contextagent,yang2025fingertip,zhao2025appagentpro,tang2026proagentbench}. Each intervention depends on two decisions: \emph{when} to interrupt and \emph{what} to ground the assistance on. Recent systems use LLMs to interpret activity histories, decide whether assistance is warranted, and prepare context for action generation \citep{lu2025proactiveagent,yang2025contextagent,xie2026pask}. This design places LLM inference on the continuous monitoring path and introduces two problems.

\begin{figure}[t]
    \centering
    \includegraphics[width=\textwidth]{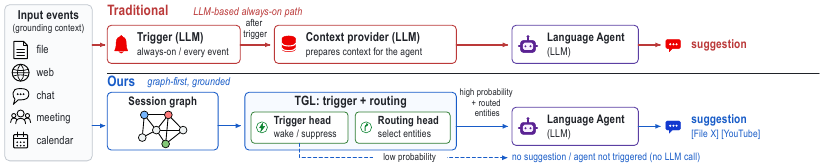}
    \caption{A shared graph controller handles triggering and context selection before language-agent generation. The upper path illustrates LLM-based implementations of these functions \citep{yang2025contextagent,xie2026pask}. \tgl{} computes trigger and entity-routing outputs in one forward pass and invokes the downstream language agent on triggered events.}
    \label{fig:intro}
\end{figure}

\noindent\textbf{Slow and expensive triggering.}
LLM latency sits directly on the path to each wake-up decision. A released desktop demo defaults to a $15$-second interval, or $240$ checks per hour; each check processes the activity history before deciding whether to offer help \citep{lu2025proactiveagent}. Even a decision to remain silent incurs this inference cost. Frequent polling trades higher resource use for faster reactions \citep{maldaner2026sentinelbench}, while keeping the model available imposes a persistent memory footprint.

\noindent\textbf{Indirect modeling of structured context.}
Existing pipelines convert structured activity records into natural-language event descriptions for LLM prediction \citep{lu2025proactiveagent}, and use LLMs to condense retrieved evidence into textual context \citep{xie2026pask}. This leaves event--entity association to language inference: repeated edits of the same script appear as separate mentions that the model must connect to identify the relevant artifact. Using relations already present in the activity records thus requires interpreting their textual expression at each decision. The descriptions also determine which cues reach the decision model: generic summaries can omit relevant details, while verbose descriptions introduce redundant information \citep{yang2025contextagent}.

Activity records associate timestamped events with entities such as files, applications, and web pages \citep{lu2025proactiveagent,yang2025fingertip}. Recurring entities connect separate events, while temporal order links successive interactions. These relations give user activity a native event--entity graph structure. Our key insight is to formulate both control decisions as node prediction on this graph: predicting whether the current event warrants assistance answers \emph{when to intervene}, and predicting which of its entities should ground that assistance answers \emph{what to anchor}.

We introduce a temporal-graph-learning (\tgl{}) controller that learns these two predictions jointly. A compact graph encoder aggregates context along event--entity and temporal connections, and trigger and routing heads use this shared representation to produce a wake-up probability and entity relevance scores in one forward pass. The controller handles continuous monitoring; on triggered events, a frozen language agent receives the activity history and scored entities to generate a suggestion (Figure~\ref{fig:intro}).

This simple design combines low-cost monitoring with stronger proactive performance (Figure~\ref{fig:teaser}). At $11.13$\,ms per event, \tgl{} achieves the highest trigger AUCs among nine architectures, exceeding the best LLM trigger by $7.0$ AUC points when ranking events that require assistance. A shared \tgl{} model improves downstream F1 across all $14$ backbones by a mean of $16.7$ points. The activity graph thus supports a fast controller that improves the language agents it guides.

Our contributions are threefold:
\begin{itemize}
    \item We formulate proactive control as joint prediction on event and entity nodes, and develop a shared temporal graph model that computes triggering and context selection in one forward pass.
    \item We show that one trained graph controller improves proactive assistance across $14$ frozen language backbones with the same threshold and prompt. Ablations measure the contributions of triggering, routed context, and the downstream prompt interface.
    \item We demonstrate deployment at $11.13$\,ms per event on a server and $13.99$\,ms on a consumer laptop, giving $\sim\!4$--$7\times$ and $\sim\!12$--$83\times$ trigger-stage speedups over the two single-forward LLM triggers. The controller has $1.16$\,M trainable parameters and a ${\sim}220$\,MiB BF16 resident footprint including its frozen text encoder.
\end{itemize}

%% file: src/2_related.tex
\section{Related Work}
\label{sec:related}

\noindent\textbf{Proactive assistance.}
Proactive assistance spans situation-aware agents, personal assistants, context-aware services, plan recognition, and recommendation \citep{earlyproactive2,myers2007proactive,hong2009context,oh2011probabilistic,woerndl2011model}. Human-computer interaction studies examine user expectations, interaction design, and satisfaction \citep{meurisch2020exploring,jones2024designing,towardshumancentered,proactive_bise2024}. LLM-based systems extend this agenda to desktop activity \citep{lu2025proactiveagent}, open-world sensory inputs \citep{yang2025contextagent,proagent}, mobile interfaces \citep{yang2025fingertip,zhao2025appagentpro,kong2026proactivemobile}, dialogue \citep{liu2025proactiveeval,zhang2025proactive}, programming, and cooperative gameplay \citep{proactive_program,proagent_gaming}. Recent methods use clarification and long-term memory to guide assistance \citep{zhang2024ask,xie2026pask}, while reinforcement learning optimizes personalized behavior and action timing \citep{sun2025ppp,ding2026proactor}. Lightweight controllers also gate reasoning and select candidate functions from multimodal inputs \citep{ding2026prpf}, or adapt intervention timing from temporal and activity preference statistics \citep{wang2026eopa}. We learn triggering and entity priorities jointly on a shared activity graph.

\noindent\textbf{Memory and personalization.}
Personalization uses user histories and preferences to adapt LLM behavior \citep{salemi-etal-2024-lamp,tan-etal-2024-democratizing,richardson2023personalization}, supported by retrieval, profiles, and accumulated experience \citep{Lewis2020,Huang2024,Shi2025,yao2022react,chen2023fireact,zhao2024expel,li2025hello,wang2025karma}. Simulated user feedback further supports learning timely, personalized recommendations \citep{kim2026propersim}. Temporal knowledge graphs organize agent memory \citep{rasmussen2025zep}; recent egocentric assistants combine graph memory, temporal summaries, and visual retrieval to ground proactive responses \citep{gong2026vinci2}. Evaluation connects these capabilities to the costs of interruption, attentional demand, and task resumption \citep{sawhney1999nomadic,gluck2007matching,iqbal2008effects,mark2008cost,adamczyk2004interruptions,iqbal2007disruption,czerwinski2004diary,parnin2013programmer}. Web, GUI, and mobile benchmarks provide interactive environments \citep{deka2017rico,deng2023mind2web,zhou2023webarena,gao2023assistgui,kapoor2024omniact,chen2024gui,rawles2024androidworld}. New proactive benchmarks assess intervention timing and content in recorded workflows \citep{tang2026proagentbench}, event tracking in ongoing conversations \citep{li2026proevent}, and response latency alongside resource use during extended monitoring \citep{maldaner2026sentinelbench}. These tasks motivate efficient decisions over evolving activity and context.

\noindent\textbf{Temporal graph learning.}
Temporal graph learning models evolving interactions through graph representations. Continuous-time and dynamic graph models such as TGAT and TGN encode time-dependent neighbourhoods \citep{xu2020inductive,rossi2020tgn}. Related work on dynamic knowledge and temporal reasoning further explored recurrent, evolutionary, and neighborhood-based graph updates \citep{DBLP:conf/icml/TrivediDWS17-know-evolve,DBLP:conf/emnlp/JinQJR20-re-net,10.1145/3404835.3462963-re-gcn,DBLP:conf/ijcai/LiS022-tirgn,DBLP:conf/acl/LiGJPL000GC22-cen}. More generally, graph encoders established practical ways to model relational structure and node-local message passing \citep{DBLP:conf/iclr/KipfW17-gcn,DBLP:conf/iclr/VelickovicCCRLB18-gat,DBLP:conf/esws/SchlichtkrullKB18-r-gcn,DBLP:conf/iclr/VashishthSNT20-compgcn}. Session graphs built from a user's interaction stream are likewise the substrate of session-based recommendation, where a GNN over one session's items predicts the next item \citep{wu2019srgnn,xu2019gcsan,wu2022gnnrecsurvey}. \tgl{} builds on both lines as a relation-aware encoder over the heterogeneous event--entity graph. We formulate proactive triggering and context selection as joint node-level prediction: event and entity readouts determine when an agent acts and which context it receives.

%% file: src/3_problem.tex
\begin{figure}[t]
    \centering
    \includegraphics[width=\textwidth]{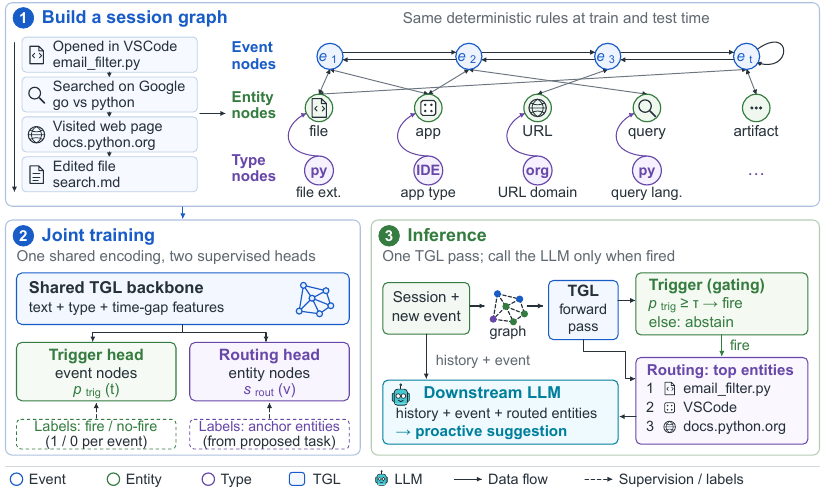}
    \caption{Joint trigger and routing prediction on the activity graph. Event, entity, and type nodes encode the session's relational structure. A shared encoder trains with event-level trigger and entity-level routing supervision. At inference, one forward pass supplies a trigger decision and scored entities to the downstream language agent.}
    \label{fig:method}
\end{figure}
\section{Problem Formulation}
\label{sec:problem}

We consider a proactive assistant that observes a timestamped event stream $\mathcal{E} = \{e_1, e_2, \ldots, e_t, \ldots\}$, where each event records a user or environment observation such as an application switch, file edit, browser visit, query, or other interaction trace. At each decision time $t$, the system must answer two coupled questions: \emph{(i)} should the downstream agent be invoked now; and \emph{(ii)} if so, which contextual entities should be routed forward as evidence?

Both decisions are produced from the observed history up to $t$, represented as a heterogeneous temporal graph $G_t = (V_t, E_t)$ over event nodes and the entities they touch.

Training provides binary event labels $y^{\mathrm{trig}}_t \in \{0,1\}$ for triggering and binary entity labels $y^{\mathrm{rout}}_v \in \{0,1\}$ for routing. These labels are derived from the released task suggestions by the procedure in Appendix~\ref{appx:graph-construction}. A shared graph encoder supports two readouts: a trigger head on event nodes and a routing head on entity nodes. At decision time $t$, the trigger score determines whether to invoke the downstream agent, and the routing scores rank the entities associated with $e_t$. Both readouts are computed in one forward pass.

The graph controller processes each event and supplies the invocation decision and selected context. The downstream LLM receives the rolling session history and routing list, then composes the suggestion. This interface gives the two stages distinct computational roles: continuous activity prediction in the graph controller and task generation in the language agent.

%% file: src/4_method.tex
\section{Method}
\label{sec:method}

We model proactive activity as a heterogeneous temporal graph $G_t = (V_t, E_t)$ (Figure~\ref{fig:method}). Nodes represent the events themselves and the semantic entities the user touches at each event (files, applications, URLs, search queries, and lightweight \emph{type nodes} that group entities by extension, topic, domain, or query language). Edges encode \texttt{has\_entity} relations from an event to its extracted entities and the reverse \texttt{belongs\_to} relation, two directional temporal edges between consecutive events (forward $t \to t{+}1$, past-informs-future; backward $t{+}1 \to t$, future-informs-past), and \texttt{self} loops. The graph is built deterministically from the event stream by a rule-based extractor, whose details (entity vocabulary, type-augmentation rules, edge channels) are given in Appendix~\ref{appx:graph-construction}.

\subsection{Joint Trigger and Routing as Two Heads}
\label{sec:method:heads}

The graph connects the two prediction tasks through their evidence. Event nodes aggregate participating entities and nearby activity, while entity nodes aggregate the events in which they appear. Joint trigger and routing supervision trains this shared encoder to support event-level wake-up decisions and entity-level context selection.

\noindent\textbf{Node features.} Every node $v$ enters the encoder as the concatenation of a frozen text embedding of its surface string, a learned embedding of its node type $\tau_v$, and, for event nodes, a learned embedding of the time gap $\Delta t_v$ since the previous event in the session (zero for entity nodes):
\begin{equation}
h^{(0)}_v = W_0\,\big[\,\mathrm{BGE}(\mathrm{text}_v)\,\|\,E_{\mathrm{type}}[\tau_v]\,\|\,\mathrm{MLP}_{\Delta t}\!\big(\log(1+\Delta t_v)\big)\,\big],
\label{eq:input}
\end{equation}
Here $\mathrm{BGE}(\cdot)$ is a frozen text encoder and $W_0$ maps the features to width $d$. During streaming inference, recurring entities reuse their cached text embeddings; each update encodes the current event and newly encountered entities. This makes persistent entity identity useful for both context aggregation and computation reuse.

\noindent\textbf{Relation-aware message passing.} Information flows over the session graph through $L{=}3$ rounds of attention-weighted message passing \citep{brody2022attentivegraphattentionnetworks}, in which every edge $(u \to v)$ carries a learned embedding $e_{r}$ of its relation $r \in \{\texttt{has\_entity}, \texttt{belongs\_to}, \texttt{fwd}, \texttt{bwd}, \texttt{self}\}$, so the attention can weight a file differently from a preceding event or from the node itself:
\begin{align}
\alpha^{(\ell)}_{vu} &= \underset{u \in \mathcal{N}(v)}{\mathrm{softmax}}\Big( a_\ell^{\top}\,\mathrm{LeakyReLU}\big( W^{\ell}_{s} h^{(\ell-1)}_v + W^{\ell}_{t} h^{(\ell-1)}_u + W^{\ell}_{e}\, e_{r(u,v)} \big) \Big), \label{eq:att}\\
h^{(\ell)}_v &= \mathrm{LN}\Big( h^{(\ell-1)}_v + \mathrm{Drop}\Big(\mathrm{GELU}\Big( \big\Vert_{k=1}^{K} \sum_{u \in \mathcal{N}(v)} \alpha^{(\ell,k)}_{vu}\, W^{\ell,k}_{t}\, h^{(\ell-1)}_u \Big)\Big) \Big), \label{eq:layer}
\end{align}
with $K$ attention heads concatenated, and a residual, LayerNorm, GELU, and dropout block around each layer. $\mathcal{N}(v)$ contains the explicit \texttt{self} edge with its own relation embedding. A Jumping Knowledge readout \citep{xu2018jumping} concatenates the input projection with every layer as $h^{\mathrm{jk}}_v = [h^{(0)}_v \,\|\, h^{(1)}_v \,\|\, h^{(2)}_v \,\|\, h^{(3)}_v]$, giving both heads access to node-local features and aggregated session context.

\noindent\textbf{Prediction heads and training objective.} The two heads are parallel MLPs on the same $h^{\mathrm{jk}}$, each applied to its own node type:
\begin{equation}
p_{\mathrm{trig}}(t) = \sigma\big(\mathrm{MLP}_{\mathrm{trig}}(h^{\mathrm{jk}}_{e_t})\big)\ \ \text{for event nodes},\qquad
s_{\mathrm{rout}}(v) = \sigma\big(\mathrm{MLP}_{\mathrm{rout}}(h^{\mathrm{jk}}_{v})\big)\ \ \text{for entity nodes},
\label{eq:heads}
\end{equation}
The joint objective combines trigger and routing losses:
\begin{equation}
\mathcal{L} = \lambda_{\mathrm{trig}}\,\mathrm{BCE}_{w_+}\!\big(p_{\mathrm{trig}}, y^{\mathrm{trig}}\big) + \lambda_{\mathrm{rout}}\,\mathrm{BCE}\big(s_{\mathrm{rout}}, y^{\mathrm{rout}}\big),
\label{eq:loss}
\end{equation}
where $w_+$ down-weights the dominant positive trigger class. Both loss terms update the shared graph encoder. Hidden width, head count, loss weights, and projection sizes are in Appendix~\ref{appx:impl}.

At test time, the trigger fires when $p_{\mathrm{trig}}(t) \geq 0.5$. The routing head ranks entities connected to the current event, and up to $20$ entries are passed to the downstream agent on triggered events. The threshold, routed-list cap, and inference procedure are shared across all downstream backbones (Appendix~\ref{appx:impl}).

\noindent\textbf{Anchor routing labels.} The released training tasks supervise both decisions. A non-null task or an explicit acceptance annotation supplies a positive trigger label. We apply the event extractor to the task text. When its entities match the event graph, the matches supply routing positives and the remaining entities at that event supply negatives. These labels connect the wake-up target to the artifacts used in the training suggestion. Appendix~\ref{appx:graph-construction} gives the complete labeling procedure for the ProactiveAgent training split, including empty-match and negative-trigger events.

\noindent\textbf{Downstream agent.} A frozen instruction-tuned chat model receives the rolling activity history and the scored entity list when the trigger fires. The list supplies concrete anchors, and the history supplies the surrounding task context. The language agent uses both to compose a user-facing suggestion in a JSON proposed-task field; with an empty list, it draws on the history alone. Prompt design and few-shot construction are described in Appendix~\ref{appx:agent}, and the same prompt serves every backbone.

%% file: src/5_experiment.tex
\section{Experiments}
\label{sec:experiment}

We evaluate two requirements of the design: a shared controller should support different language agents and remain inexpensive to run continuously. We first compare suggestion quality across backbones, then examine invocation timing and routed context through pipeline ablations. Trigger-architecture and deployment comparisons assess the quality and cost of the always-on decision.

\subsection{Setup}
\label{sec:experiment:setup}

We evaluate proactive triggering and routing on the desktop ProactiveAgent benchmark using its released splits and reward-model judge \citep{lu2025proactiveagent}. The judge scores generated task suggestions under the benchmark's precision, recall, accuracy, false-alarm, and F1 definitions. We use F1 to assess the complete suggestion pipeline and precision and recall to distinguish false alarms from missed opportunities. Vanilla results use the originating paper's prompt protocol; \emph{+\,Ours} supplies the frozen downstream language agent with the graph controller's trigger and routed context. Appendix~\ref{appx:fingertip} evaluates the graph-encoder design on mobile intent prediction with FingerTip-20K \citep{yang2025fingertip}.

We use the same \tgl{} model, trigger threshold, and prompt template across all backbones. We evaluate $14$ instruction-following backbones spanning four families (open-weight, OpenAI, Anthropic, DeepSeek) as the downstream language agent, with two fine-tuned proactive baselines from \citet{lu2025proactiveagent} for reference. Backbone list, model architecture, prompt, serving setup, and graph-construction rules are deferred to Appendices~\ref{appx:graph-construction},~\ref{appx:agent}, and~\ref{appx:impl}.

\input{src/6_results}

\input{src/7_efficiency}

%% file: src/6_results.tex
\subsection{Main Results}
\label{sec:main-results}

\begin{figure}[t]
    \centering
    \includegraphics[width=\textwidth]{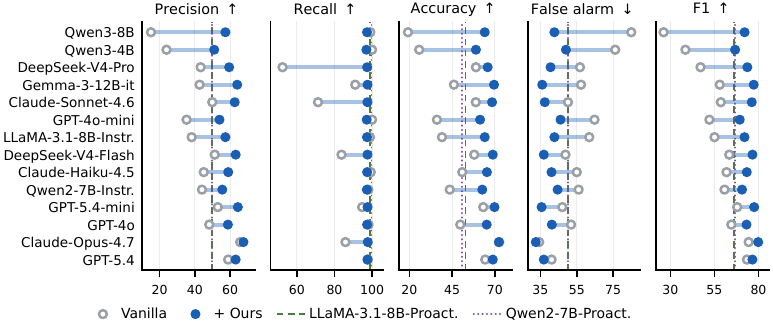}
    \caption{One graph controller improves F1 across all $14$ downstream backbones. Open and filled markers show vanilla and \emph{+\,Ours}; vertical reference lines show the two published fine-tuned proactive baselines. The mean F1 gain is $16.7$ points; on the two corresponding backbones, \emph{+\,Ours} also exceeds the published fine-tuned baselines. All backbones use the same \tgl{} model and threshold. Vanilla values combine published legacy results and our recent-model runs; Table~\ref{tab:main-results} gives their provenance and exact values.}
    \label{fig:main-results}
\end{figure}

The controller selects intervention times and entities independently of the downstream backbone; the language agent determines the resulting suggestion. We therefore evaluate the same controller with $14$ backbones to test its usefulness across different generators. Figure~\ref{fig:main-results} compares each backbone with \tgl{} against its vanilla configuration under the protocol of \citet{lu2025proactiveagent}. Appendix~\ref{appx:full-main-results} gives every metric and identifies the published and re-run vanilla values.

\noindent\textbf{A shared controller improves every backbone.} \tgl{} raises F1 on all $14$ downstream backbones by $3.1$--$46.0$ points, with a mean gain of $16.7$. Gains span open-weight models, including Qwen3-8B ($+46.0$) and LLaMA-3.1-8B ($+17.0$), and API models, including DeepSeek-V4-Pro ($+26.6$), Claude-Sonnet-4.6 ($+17.6$), and GPT-5.4 ($+3.1$). Every \emph{+\,Ours} configuration uses the same trigger model, threshold, routing interface, and prompt. The controller can therefore be trained once and reused to improve different language agents, retaining the same control decisions as the downstream backbone changes.

\noindent\textbf{Comparison with proactive fine-tuning.} Proactive behavior can be learned by adapting the language agent or training a separate controller. We compare these choices on corresponding backbones to examine where to place task-specific training. The proactive models published by \citet{lu2025proactiveagent}, LLaMA-3.1-8B-Proactive and Qwen2-7B-Proactive, reach F1 scores of $66.25$ and $66.47$. On the corresponding frozen instruction-tuned backbones, \emph{+\,Ours} reaches $72.07$ and $70.68$, respectively. Training the $1.16$\,M-parameter controller thus achieves higher F1 than proactive fine-tuning of these two LLMs. Appendix~\ref{appx:dropin} further applies the trigger to the fine-tuned agents themselves: filtering their existing suggestions improves F1 for both, showing that the controller also complements prior proactive adaptation.

\noindent\textbf{Higher precision with high recall.} We examine precision and recall separately to characterize the F1 gains under the benchmark's evaluation protocol. Qwen3-8B starts with $99\%$ recall and $15\%$ precision; \tgl{} raises its precision to $57\%$ at $97\%$ recall. DeepSeek-V4-Pro and Claude-Sonnet-4.6 start with recall of $52\%$ and $71\%$; both exceed $97\%$ with \tgl{}. Across all backbones, precision improves and benchmark recall reaches at least $96\%$.

\subsection{Ablation Study}
\label{sec:results:anatomy}

\begin{figure}[t]
    \centering
    \includegraphics[width=\textwidth]{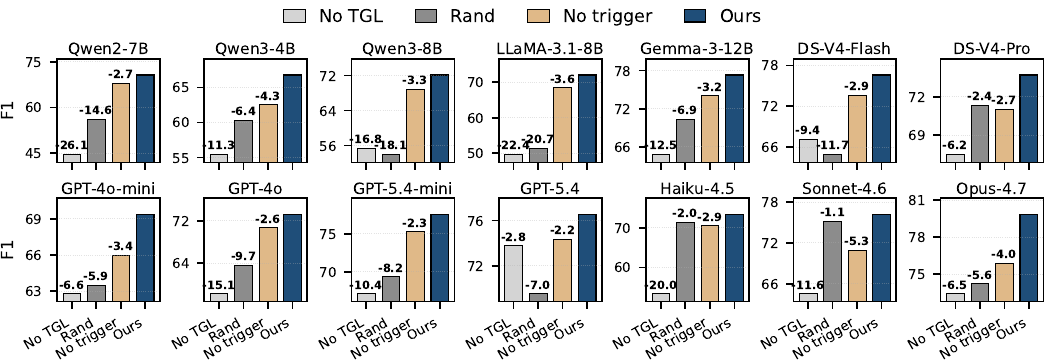}
    \caption{\tgl{} improves F1 over all three controls on every backbone. Each panel compares \emph{Ours} with prompt-only self-gating (No \tgl{}), random routing (Rand), and generation at every event (\mbox{No trigger}).}
    \label{fig:anatomy}
\end{figure}

We next examine the controller's two responsibilities: selecting intervention times and supplying entity context. The first two controls change invocation timing or routed content within the controller-guided pipeline; the third compares the complete interface with prompt-only self-gating. Figure~\ref{fig:anatomy} evaluates all three on the same $14$ backbones. Appendix~\ref{appx:anatomy-full} reports every per-backbone result.

\noindent\textbf{Does the trigger matter?} We first test the value of deciding when to invoke an agent that already receives routed context. \emph{No trigger} invokes it on every event with the same routing list and prompt. Restoring the trigger improves mean F1 by $3.2$ points, with gains of $2.2$--$5.3$ across all $14$ backbones. Selecting intervention times therefore improves suggestion quality, and the skipped events also save downstream calls (Section~\ref{sec:efficiency}).

\noindent\textbf{Does the routing content matter?} The agent receives both session history and routed entities. To examine how the supplied entities affect its suggestions alongside that history, \emph{Rand.\ routes} replaces each routing list with an equal-length list sampled from the training entity vocabulary. The history, trigger, and downstream prompt remain fixed. Using the learned routing list improves mean F1 by $8.6$ points, with gains of $1.1$--$20.7$ across all $14$ backbones. The largest differences are on LLaMA-3.1-8B ($20.7$), Qwen3-8B ($18.1$), and Qwen2-7B ($14.6$). Graph-selected anchors consistently outperform unrelated entities supplied through the same interface.

\noindent\textbf{Comparison with prompt-only self-gating.} Output-style instructions also guide task formulation. We therefore compare the complete controller interface with an LLM that decides when and what to suggest under the same style requirements. The prompt-only baseline \emph{No \tgl{}} retains the output-style directives: a single sentence, at most $150$ characters, led by a concrete deliverable. It asks the agent to choose between a task and an empty proposed-task string, omitting the routing field, routing-anchor instructions, and three routing demonstrations. \emph{Ours} improves mean F1 by $12.7$ points, with gains of $2.8$--$26.1$ across all backbones. The controller-guided interface produces better suggestions than language-agent self-gating under the same output-style requirements.

\subsection{Trigger Architecture Comparison}
\label{sec:results:trigger-arch}

\input{src/tables/trigger_arch_ablation}

Choosing an always-on trigger requires balancing decision quality with the cost of repeated evaluation. Table~\ref{tab:trigger-arch} compares nine architectures across five families---Rule, Tabular, Textual, LLM, and Graph---on trigger ranking and inference cost. Rule and tabular models test prediction from inexpensive event cues. Text classifiers test semantic encoding with the same BGE encoder used by \tgl{}, including frozen and fine-tuned variants to examine the effect of encoder adaptation. The LLM rows cover both direct binary prediction and deriving a decision from task generation.

LR and HGB use event features; BGE-frozen + MLP and BGE-FT + MLP encode the latest five event texts; the \tgl{} encoder processes the event--entity session graph. The LLM family includes fine-tuned Qwen3-$0.6$B and Qwen3-$8$B with a yes/no logit readout, plus Proactive-Qwen3-$0.6$B with a trigger derived from its generated task. Training data come from the same ProactiveAgent release, with event-label supervision for trigger classifiers and an additional entity-label objective for \tgl{}. For downstream evaluation, each trigger supplies its decision to the same \tgl{} routing list, prompt, and language agents. Appendix~\ref{appx:trigger-baselines} specifies the inputs and training recipe for each architecture.

\noindent\textbf{Trigger evaluation labels.} A trigger must recognise opportunities for assistance and occasions to remain silent. We assess their ranking with three binary labelings of the released test annotations. The \emph{must-fire} set contains events requiring a non-null task (${\sim}24\%$). The \emph{can-skip} set contains events accepting an empty suggestion (${\sim}51\%$); its \emph{must-skip} subset contains events rejecting every alternative suggestion (${\sim}14\%$ of test events). The three AUC columns use, respectively, membership in must-fire, membership outside can-skip, and membership outside must-skip as the positive label. Each column evaluates a separate target, with higher scores indicating a preference to fire.

\noindent\textbf{\tgl{} leads every trigger AUC.} \tgl{} reaches $0.738$ AUC on must-fire ranking, $+7.0$ pp above the strongest LLM trigger (Qwen3-0.6B at $0.668$). The two skip-side columns show the same pattern: \tgl{} at $0.658$ on can-skip and $0.639$ on must-skip, $+4.0$ and $+4.2$ pp above the next best (Qwen3-8B at $0.618$ and $0.597$). The AUC lead comes with lower latency: \tgl{} runs at $11.13$\,ms per event versus $40.4$--$3{,}927$\,ms for the three LLM-as-trigger configurations on the same server. The graph controller combines better ranking of fire and skip decisions with lower monitoring cost.

\noindent\textbf{Deciding before generation reduces monitoring cost.} The LLM rows also expose the cost of deriving a trigger from a generated task. The two LoRA Qwen3 triggers produce a yes/no score in one forward pass, costing $40.4$ and $78.6$\,ms per event on the server. Proactive-Qwen3-$0.6$B generates a task string and maps a non-empty task to fire. Its mean $139$ generated tokens cost $3{,}927$\,ms on the server and $11{,}966$\,ms on the laptop; the corresponding trigger AUCs are $0.505$, $0.508$, and $0.485$. Predicting the intervention decision before composing a task removes decoding from the continuous monitoring path. \tgl{} implements this separation with a single forward pass that also supplies the routing list.

\noindent\textbf{A shared threshold supports backbone changes.} Reusing the controller across language agents also raises the question of whether its operating threshold needs adjustment. We use $\tau=0.5$ for all $14$ backbones and evaluate both sensitivity to this setting and transfer between backbones. On five open-weight backbones with three runs each, mean F1 ranges from $70.4$ to $73.1$ over $\tau\in[0.4,0.6]$, with $72.1$ at the deployed threshold (Appendix~\ref{appx:rebuttal:threshold}). We also choose a threshold on four backbones and transfer it to the fifth. This incurs a mean loss of $0.4$ F1 relative to that backbone's tuned threshold. The controller can therefore reuse its operating threshold across these backbones with little loss in F1. Table~\ref{tab:trigger-arch} additionally reports Brier score, ECE, and the across-backbone spread of F1-maximising thresholds; Appendix~\ref{appx:rebuttal:threshold} reports temperature calibration, and Figure~\ref{fig:score-distributions} shows the score distributions.

%% file: src/tables/trigger_arch_ablation.tex
\begin{table}[t]
\centering
\small
\setlength{\tabcolsep}{2.6pt}
\renewcommand{\arraystretch}{1.15}
\setlength{\abovecaptionskip}{0pt}
\caption{\tgl{} achieves the highest trigger AUCs with lower latency than the LLM triggers. Architectures are trained from the ProactiveAgent release. AUC uses the three binary labelings in Section~\ref{sec:results:trigger-arch}; Brier and ECE use the calibration target described in Appendix~\ref{appx:rebuttal:threshold}. Trig.\ std describes the spread of F1-maximising thresholds across backbones. LLM latency includes single-forward and generation-derived protocols.}
\label{tab:trigger-arch}
\begin{tabular}{l | l | rrr | rr | r | rr}
\toprule
                                 &         & \multicolumn{3}{c|}{\textbf{Trigger AUC} $\uparrow$}                                & \multicolumn{2}{c|}{\textbf{Calibration} $\downarrow$} & \multicolumn{1}{c|}{\textbf{Thresholds}} & \multicolumn{2}{c}{\textbf{Latency} (ms) $\downarrow$} \\
\cmidrule(lr){3-5} \cmidrule(lr){6-7} \cmidrule(lr){8-8} \cmidrule(lr){9-10}
Trigger                          & Type    & m-fire & c-skip & m-skip & Brier  & ECE    & Trig.\ std & Server & Local \\
\midrule
Rule                             & Rule    & $0.585$               & $0.581$               & $0.496$               & $0.319$ & $0.294$ & $0.143$    & $0.0003$    & $0.0001$   \\
\midrule
LR ($65$ features)               & Tabular & $0.580$               & $0.534$               & $0.499$               & $0.270$ & $0.201$ & $0.115$    & $0.13$      & $0.03$     \\
HGB ($65$ features)              & Tabular & $0.636$               & $0.551$               & $0.500$               & $0.309$ & $0.267$ & $0.086$    & $7.29$      & $5.73$     \\
\midrule
BGE-frozen + MLP                 & Textual & $0.574$               & $0.518$               & $0.460$               & $0.601$ & $0.611$ & $0.358$    & $6.59$      & $11.95$    \\
BGE-FT + MLP                     & Textual & $0.621$               & $0.586$               & $0.553$               & $0.593$ & $0.599$ & $0.396$    & $6.86$      & $11.06$    \\
\midrule
Qwen3-0.6B                       & LLM     & $0.668$               & $0.582$               & $0.504$               & $0.584$ & $0.596$ & $0.259$    & $40.4$      & $162.3$    \\
Qwen3-8B                         & LLM     & $0.644$               & $0.618$               & $0.597$               & $0.580$ & $0.592$ & $0.222$    & $78.6$      & $1156.8$   \\
Proactive-Qwen3-0.6B (SFT)  & LLM     & $0.505$               & $0.508$               & $0.485$               & $0.657$ & $0.657$ & $0.140$    & $3{,}927$   & $11{,}966$ \\
\midrule
\tgl{} (Ours)                    & Graph   & $0.738$               & $0.658$               & $0.639$               & $0.308$ & $0.334$ & $0.035$    & $11.13$     & $13.99$    \\
\bottomrule
\end{tabular}
\end{table}

%% file: src/7_efficiency.tex
\subsection{Efficiency}
\label{sec:efficiency}

The controller runs at every observation, including those that produce no suggestion, and keeps its weights resident between checks. We therefore measure both per-event latency and the full memory footprint, including the text encoder. An NVIDIA RTX A6000 (48\,GB) and an Apple M3 Pro (36\,GB) cover server and consumer-laptop deployment (Appendix~\ref{appx:latency-protocol}).

\noindent\textbf{Per-event latency.} \tgl{} runs at $11.13$\,ms on the server. BGE encoding takes ${\sim}7$--$8$\,ms for the event text and a fresh entity encoded together, and the graph layers add ${\sim}3$\,ms. Recurring entities reuse cached embeddings, allowing the graph to retain their context with little additional encoding work. The Qwen3-$0.6$B and Qwen3-$8$B single-forward triggers take $40.4$ and $78.6$\,ms, giving \tgl{} a $\sim\!4$--$7\times$ trigger-stage speedup.

\noindent\textbf{Consumer-laptop deployment.} The same \tgl{} model runs at $13.99$\,ms per event on the M3 Pro. Qwen3-$0.6$B and Qwen3-$8$B take $162.3$ and $1{,}156.8$\,ms, respectively, giving trigger-stage speedups of ${\sim}12\times$ and ${\sim}83\times$. The Tabular, Textual, and Graph triggers all run within $14$\,ms on this laptop. \tgl{} combines the highest trigger AUC with sub-$14$\,ms inference on this consumer device.

\noindent\textbf{Memory footprint.} \tgl{} has $1.16$\,M trainable parameters and a $109$\,M frozen BGE encoder. BF16 resident weights occupy ${\sim}220$\,MiB, and streaming inference peaks at ${\sim}267$\,MiB. The Qwen3-$0.6$B and Qwen3-$8$B triggers require approximately $1.5$\,GB and $16$\,GB of BF16 weights, or ${\sim}7$--$75\times$ the controller's resident footprint (Table~\ref{tab:trigger-resources}). The controller, including its text encoder, occupies a small share of laptop memory and leaves room for the user's applications.

\noindent\textbf{Downstream generation.} Graph prediction determines the cost of each monitoring check; triggered events additionally incur a language-agent call. Appendix~\ref{appx:per-backbone-latency} reports the wall-clock cost of each backbone's generation call. The \emph{No trigger} ablation in Section~\ref{sec:results:anatomy} evaluates the effect of invocation timing on suggestion quality.

%% file: src/10_conclusion.tex
\section{Conclusion}
\label{sec:conclusion}

Continuous proactive assistance requires inexpensive decisions about when to intervene and which context should guide the intervention. We have shown that user activity's native event--entity structure supports both decisions through a small graph model. \tgl{} links events through shared entities and jointly learns event-level triggering and entity-level context selection, producing both outputs in one forward pass. It achieves the highest AUCs among nine trigger architectures with $\sim\!4$--$7\times$ trigger-stage speedups over the two single-forward LLM triggers. A shared \tgl{} model improves F1 across all $14$ downstream backbones by a mean of $16.7$ points and runs at $13.99$\,ms per event on a consumer laptop. These results establish a practical design for proactive agents: a compact graph model handles continuous monitoring and context selection, and a language agent turns the selected context into user-facing assistance.

%% file: src/K_ethics.tex
\section*{Ethics Statement}

This work uses the publicly released ProactiveAgent and FingerTip-20K benchmarks under their original terms \citep{lu2025proactiveagent,yang2025fingertip}. The experiments use the released activity records, and our code release contains the graph-construction and modelling pipeline. The study develops an efficient trigger and context-selection controller for proactive agents.

Always-on activity logs are privacy-sensitive: file names, URLs, search queries, application switches, and (on FingerTip) demographic profile fields all flow through the trigger and routing heads. The graph controller's small footprint supports on-device activity processing (Section~\ref{sec:efficiency}). Deployment must also account for the activity context sent to the downstream language agent. Production deployment additionally requires standard data minimisation, sensitive-entity filtering, opt-out, and retention controls. Shared-device deployments require separate per-user graphs and access controls. A badly-timed suggestion interrupts focused work and repeated mistimed interventions erode trust. The trigger head is trained to suppress such events, and a deployment should monitor this failure mode directly. Fairness analysis across the FingerTip demographic subgroups (gender, age, occupation, location, device) is required before any are used as routing features in a deployed system.

%% file: src/N_reproducibility.tex
\section*{Reproducibility Statement}

We evaluate on ProactiveAgent with its released reward-model judge \citep{lu2025proactiveagent} and on FingerTip-20K with its released splits and metrics \citep{yang2025fingertip}. Table~\ref{tab:main-results} identifies published reference values and our evaluation runs.

Appendix~\ref{appx:graph-construction} specifies graph extraction, node types, edge relations, and label derivation. Appendix~\ref{appx:impl} gives the encoder, heads, training settings, inference procedure, sampling settings, and latency protocol. The downstream system prompts appear in Appendix~\ref{appx:prompts}. Appendix~\ref{appx:rebuttal} reports retraining variation, threshold sensitivity, and construction variants. Code and the graph-construction pipeline will be released.

%% file: src/L_llm_use.tex
\section*{AI Use Statement}

We use LLMs as general research assistants for literature discovery, feedback on experimental design, code implementation and debugging, result analysis, and manuscript drafting and editing. The authors review AI-assisted outputs and take responsibility for the paper, code, and reported results.

%% file: src/A_fingertip.tex
\section{Graph-Based Mobile Intent Prediction}
\label{appx:fingertip}

To examine graph-based context selection for mobile intent prediction, we apply the temporal graph-encoder design to FingerTip-20K \citep{yang2025fingertip}. Each episode supplies a graph of user context, historical activity, and categorical entities. We train the \tgl{} encoder from scratch for $40$ epochs on FingerTip's training split and use its embedding of the current episode to retrieve training intents. We retain candidates for the application named by the highest-ranked retrieved intent before passing them to the downstream \texttt{Qwen2.5-VL-7B-Instruct} agent (Appendices~\ref{appx:graph:fingertip} and \ref{appx:agent:fingertip}).

\subsection{Evaluation Protocol}

We adopt the same evaluation as FingerTip-20K. The proactive task-suggestion track formalises the task as $I = f(U, T, S, I_{\mathrm{history}}, O)$, where the agent predicts a Chinese natural-language intent $I$ from the user profile $U$, the timestamp $T$, the scenario (location category) $S$, the user's last $20$ historical intents $I_{\mathrm{history}}$ drawn from the global usage log, and $t$ initial screenshots $O$ from the current episode ($t \in \{0,1,2,3\}$ denotes the number of screenshots). We evaluate on the public \texttt{Test-suggestion} split and report two metrics: intent similarity, $\mathrm{Sim}_1 = (S_1 + S_2)/2$, where $S_1$ is the \texttt{paraphrase-multilingual-}\allowbreak\texttt{MiniLM-L12-v2} cosine similarity and $S_2$ is the Levenshtein similarity, both normalised to $[0,1]$; and intent success rate $\mathrm{SR}_1$, the percentage of predicted intents judged equivalent to their targets by DeepSeek-V3.

\subsection{Results}

\input{src/tables/fingertip_results}

Table~\ref{tab:fingertip} compares against the published numbers from \citet{yang2025fingertip}. \tgl{} plus Qwen2.5-VL-7B with two screenshots exceeds the paper's strongest model fine-tuned with low-rank adaptation (LoRA) on both metrics ($\mathrm{SR}_1 = 26.5\%$ vs.\ $26.0\%$, $\mathrm{Sim}_1 = 0.61$ vs.\ $0.55$). With three screenshots, our system exceeds the published GPT-4.1 baseline using three screenshots and full history: we improve $\mathrm{Sim}_1$ by $+0.04$ and $\mathrm{SR}_1$ by $+12$ percentage points using only the last $20$ history events. The graph-based retrieval pipeline provides contextual intent candidates to the frozen Qwen2.5-VL-7B agent.

%% file: src/tables/fingertip_results.tex
\begin{table}[htbp]
\centering
\small
\setlength{\tabcolsep}{6pt}
\setlength{\abovecaptionskip}{0pt}
\caption{Mobile intent prediction on FingerTip-20K's public \texttt{Test-suggestion} split. Intent similarity ($\mathrm{Sim}_1$) averages embedding and edit-distance similarity; success rate ($\mathrm{SR}_1$) is judge-assessed intent equivalence. The first five rows are published baselines from \citet{yang2025fingertip}. Our rows use frozen Qwen2.5-VL-7B and the last $20$ historical intents. LoRA denotes low-rank adaptation; rank is the adapter dimension.}
\label{tab:fingertip}
\begin{tabular}{l rr}
\toprule
Method and image input & \shortstack{Intent similarity\\$\mathrm{Sim}_1\uparrow$} & \shortstack{Success rate\\$\mathrm{SR}_1$ (\%) $\uparrow$} \\
\midrule
Qwen2.5-VL-7B (0 screenshots)              & 0.25 &  3.1 \\
GPT-4.1 (0 screenshots)                    & 0.35 &  7.2 \\
GPT-4.1 (3 screenshots, full history)        & 0.55 & 13.8 \\
Qwen2.5-VL-7B fine-tuned (LoRA rank 4)               & 0.52 & 20.3 \\
Qwen2.5-VL-7B fine-tuned (LoRA rank 64)              & 0.55 & 26.0 \\
\midrule
\textbf{\tgl{} + Qwen2.5-VL-7B (2 screenshots)} & \textbf{0.61} & \textbf{26.5} \\
\textbf{\tgl{} + Qwen2.5-VL-7B (3 screenshots)} & 0.59 & 26.2 \\
\bottomrule
\end{tabular}
\end{table}

%% file: src/B_graph_construction.tex
\section{Per-Session Graph Construction}
\label{appx:graph-construction}

\paragraph{Activity structure and graph schema.} User activity links events to persistent entities such as applications, files, web addresses, and search queries. Our graph schema represents these entities as typed nodes, connects them to the events that involve them, and records event order through temporal edges. Platform metadata supplies application names, file types, URL domains, and timestamps. The benchmark implementations map their released activity records into this schema.

\paragraph{Shared representation.} Each graph contains activity nodes, entity nodes, and temporal edges linking activities in order. Desktop activity nodes represent individual events. A mobile graph contains historical activity nodes and one \emph{current-episode node}, whose representation predicts the user's next intent. Entity nodes represent named objects such as applications and files. Additional \emph{category nodes} represent properties shared by many objects, such as a file extension or an intent class. Every node has a frozen text embedding, a learned node-type embedding, and an embedding of its time offset. The three-layer graph neural network (GNN) combines these features through message passing (Appendix~\ref{appx:impl}). Each platform supplies its own node and relation vocabulary and prediction heads.

\paragraph{Entity extraction.} The desktop implementation applies regex patterns and categorical mappings to ProactiveAgent's released event descriptions. ProactiveAgent constructs these descriptions from collected and simulated activity, including language-model conversion of structured records to text \citep{lu2025proactiveagent}. The extractor identifies applications, files, web addresses, search queries, and quoted artifacts, and the full event text supplies the event-node embedding. FingerTip provides categorical episode fields that map directly to typed nodes. These two input formats share the event--entity graph representation described below.

\subsection{Instantiation: Desktop Activity Logger (ProactiveAgent)}
\label{appx:graph:desktop}

We construct each desktop session graph with deterministic entity extraction, category-node construction, and temporal linking.

\subsubsection*{Inputs}

For each session we receive a sequence of timestamped event strings from the released ProactiveAgent data. Examples: ``\emph{The user opened `email\_filter.py' in Visual Studio Code.}'', ``\emph{The user searched for `js fade-in' on Bing.}''. Each session forms an independent graph. Metadata-only events (lines starting with ``\texttt{\# Assistant Available Operations}'' or containing ``\texttt{Operation(name=}'') are filtered out before extraction.

\subsubsection*{Entity Extraction}

The extractor identifies five entity classes and adds shared category nodes:

\emph{(i)} \textbf{Applications.} Alias lookup maps names to eight canonical application labels: Visual Studio Code, IntelliJ IDEA, Outlook, Jupyter Notebook, Terminal, Browser, Email Client, and IDE. For example, Visual Studio Code, VSCode, and Code.exe map to the same application node; Chrome, Bing, and Google map to Browser.

\emph{(ii)} \textbf{Files.} A recognised filename extension identifies a file. Each file also contributes two category nodes: its \emph{file extension} (e.g., Python's \texttt{.py}) and its \emph{file category} (code, writing, data, slides, or other). The extension inventory contains $25$ extensions observed in the training stream and grouped using standard file-type conventions.

\emph{(iii)} \textbf{Web addresses (URLs).} A string beginning with \texttt{http://} or \texttt{https://} identifies a web address. Its \emph{website-domain category} is obtained by stripping a leading \texttt{www.} and retaining the final two dot-separated hostname components.

\emph{(iv)} \textbf{Search queries.} Search and typing patterns, such as ``searched for \ldots'' or ``typed \ldots'', identify the query text. A shared \emph{query-language category} groups the text as Chinese, English, mixed, or other using its Unicode script. This heuristic assigns romanised Chinese written in Latin letters to the English category.

\emph{(v)} \textbf{Document artifacts.} A quoted title or name is extracted when preceded by a document noun: file, document, report, spreadsheet, article, tab, summary, email, or draft. Matches are merged with filename matches, so ``document `report.docx''' yields one file node.

Application aliases and extraction patterns are fixed from the training stream. Entities with the same type and lowercased surface label share a node within a session. Stable identifiers combine the entity type with the first $16$ hexadecimal characters of the label's SHA-1 hash. Graph instances are created separately for each session. The code supplies the full application-alias and extension-to-category mappings.

Table~\ref{tab:rule-inventory} summarises the rules. For example, ``The user opened `email\_filter.py' in Visual Studio Code'' contributes two named entities---the file and application---and two category nodes---Python extension and code file. Both named entities and category nodes are connected to the event and are eligible for routing. Appendix~\ref{appx:rebuttal:construction} tests the effect of removing these node groups.

\begin{table}[htbp]
\centering
\small
\setlength{\tabcolsep}{6pt}
\setlength{\abovecaptionskip}{0pt}
\caption{Desktop graph construction. Named entities are extracted from event text; shared category nodes encode their file, website, or language properties.}
\label{tab:rule-inventory}
\begin{tabular}{p{0.21\linewidth}p{0.34\linewidth}p{0.34\linewidth}}
\toprule
Entity class & Extraction rule & Added category nodes \\
\midrule
Application & Application-name aliases & -- \\
File & Recognised filename extension & File extension; file category \\
Web address & Web-address syntax & Website domain \\
Search query & Search or typing pattern & Query-language category \\
Document artifact & Quoted name after a document noun & -- \\
\bottomrule
\end{tabular}
\end{table}

\subsubsection*{Event Records and Encoder Edges}

The extraction record stores each event, its entities, and their order. It marks the first application mentioned in an event as its \emph{primary entity}, falling back to the first extracted entity when no application is present. Links between consecutive primary entities preserve interaction provenance in the extraction record. The encoder reads five relation channels: event-to-entity, entity-to-event, forward temporal, backward temporal, and self-loop. Forward temporal edges connect an earlier event to the next event; backward edges connect the later event to the earlier one. Each relation has a learned embedding.

Training uses both temporal directions. At inference, backward temporal edges are removed and future-node input representations are masked according to Appendix~\ref{appx:causal}. Appendix~\ref{appx:rebuttal:causal} evaluates this training choice. Each event node records the elapsed time since its predecessor; entity and category nodes have time offset $\Delta t=0$. The routing head ranks the entity and category nodes connected to the current event. Its highest-scoring node is the \emph{routing anchor} supplied to the language agent.

\subsubsection*{Trigger and Entity-Relevance Labels}

Each session in the released ProactiveAgent training set \citep{lu2025proactiveagent} comes with a (\texttt{history}, \texttt{current\_observation}, \texttt{proposed\_task}) triple where \texttt{proposed\_task} is either a non-null task suggestion or \texttt{null}. We derive supervision as follows.

\textbf{Trigger label} on the event node: $1$ iff the assistant proposed a non-null task at this event, else $0$. The benchmark also marks acceptance in-line via the substring ``\texttt{Is Accepted: True}'', which forces label $1$ on the corresponding event even when the assistant turn is null at that turn.

\textbf{Routing labels} on entity nodes. We apply the same extractor to the proposed-task text and match its surface forms to entities in the event graph. Matches supply positive routing labels and the other entities at that event supply negative labels. A positive-trigger event with an empty match set contributes trigger supervision alone. Negative-trigger events supply negative labels for their associated entities. The session-graph loader assigns one training label to each entity node; if that entity has multiple label-table entries within a session, the last entry supplies its training target. At inference, entity scores are recomputed for the current event history and the routing list is restricted to nodes connected to that event.

The comparison uses the same released training-supervision source across the learned trigger rows and the proactive-finetuning baselines, while evaluation inputs are the held-out event histories.

\subsubsection*{Train/Test Isolation}

ProactiveAgent provides separate training and evaluation sessions. Category-level entities, including application names and file-type nodes, recur across sessions and splits. The graph representation also includes session-specific file names, search queries, and document artifacts. Each session is processed independently. An event-text matching check found that $0\%$ of evaluation event strings also occur in the training set.

\subsection{Instantiation: Mobile Usage Logger (FingerTip)}
\label{appx:graph:fingertip}

FingerTip provides application, intent-class, scenario, and time fields for each episode, along with a user profile containing gender, age, occupation, location, and device. Applications use Android package identifiers; a name lookup built from training intents supplies Chinese labels such as 微信 (WeChat). Intent classes follow the benchmark's $42$-class taxonomy. Each episode yields one graph with seven node types and eight edge relations.

\subsubsection*{Node Types}

\emph{(i)} \textbf{Current episode:} one node containing the current time, scenario, and user profile. This node's representation is used for prediction. Its Chinese text concatenates these fields with the labels 时间 (time), 场景 (scenario), and 用户画像 (user profile).

\emph{(ii)} \textbf{Historical activity:} one node for each of the user's last $20$ intents in the released usage log. Its text contains that activity's time, scenario, application, intent class, and intent description, introduced by the Chinese labels 时间, 场景, App, 类, and 意图.

\emph{(iii)} \textbf{Application:} one node per distinct application in the history, using the Chinese name obtained from the training-set lookup.

\emph{(iv)} \textbf{Intent class:} one node per distinct class in the history, such as alarm/timer, video, or shopping.

\emph{(v)} \textbf{Scenario:} one node per location category, such as home, workplace, or transit.

\emph{(vi)} \textbf{Hour of day:} one node per distinct hour, retaining the hour in its text and adding one of five Chinese time-of-day labels: late night ($0$--$5$), morning ($6$--$11$), noon ($12$--$13$), afternoon ($14$--$17$), or evening ($18$--$23$).

\emph{(vii)} \textbf{User profile:} one node containing the profile supplied for the user. Entity and category nodes are reused within the episode graph when their type and value match.

\subsubsection*{Edge Relations}

Five entity relations connect activity nodes to applications, intent classes, scenarios, hours, and user profiles. The current-episode node connects to its profile, scenario, and hour. Each historical-activity node connects to its application, intent class, scenario, and hour. These entity links carry the same relation label in both directions. Two temporal relations connect consecutive historical activities: earlier-to-later and later-to-earlier. They also link the most recent activity to the current episode in both directions. A self-loop relation lets every node retain its own features during attention. The same history graph is used during training and inference.

\subsubsection*{Heads and Training}

The current-episode node's final representation drives the prediction. Three heads sit on it: a $462$-class app classifier, a $42$-class intent-class classifier, and a $1024$-d semantic-alignment head whose target is the BGE embedding of the target intent text; the semantic head supplies the retrieval vector at inference, and the two classifiers provide auxiliary training losses. The text encoder is \texttt{BAAI/bge-large-zh-v1.5} ($1024$d) for Chinese text. The model is trained for $40$ epochs on FingerTip's released training split with AdamW (learning rate $3 \times 10^{-4}$, weight decay $10^{-4}$), batch $32$ graphs. Total trainable parameters are $\approx 2.5$\,M. These heads implement FingerTip's episode-level intent prediction task.

%% file: src/C_downstream_agent.tex
\section{Downstream Language Agent: Design Details}
\label{appx:agent}

\paragraph{Shared design.} A frozen instruction-tuned language model receives structured context and few-shot demonstrations, then emits a JSON object containing the proactive suggestion. The desktop implementation serves all $14$ downstream backbones in Table~\ref{tab:main-results}. Local vLLM inference uses temperature $0$, and API-served backbones use their default sampling settings.

\paragraph{Per-platform inputs.} The desktop agent receives the rolling session history, current observation, and scored routing entities. The FingerTip agent receives the user profile, scenario, time, latest $20$ historical intents, retrieved candidates, and episode screenshots.

\paragraph{Output formats.} We generate an English task suggestion for ProactiveAgent and a Chinese intent string for FingerTip. The following subsections specify our prompts, length limits, decoding settings, and parsing procedures.

\subsection{Instantiation: Desktop Agent (ProactiveAgent)}
\label{appx:agent:desktop}

The desktop language agent generates a task suggestion from the current observation, recent history, and controller output.

\subsubsection*{Inputs and Output Schema}

For each event the agent receives three inputs: \emph{(i)} the rolling session history, the last five raw event texts in the current session (appended at every event regardless of whether the agent fires, so non-fire events still contribute context to subsequent fire events); \emph{(ii)} the current observation, the event text at step $t$ describing what the user is doing right now; and \emph{(iii)} the routed nodes, the highest-scoring entities from the \tgl{} routing head with their scores, e.g.\ \texttt{[\{"node": "email\_filter.py",} \texttt{"score": 0.78\}, ...]}, sorted by descending $s_{\mathrm{rout}}$.

The routing list provides named entities and their relevance scores. The output JSON contains a proposed-task string and five auxiliary fields: session theme, purpose, thoughts, response, and intent. The proposed-task field is evaluated by the benchmark; the auxiliary fields are logged. An empty task string represents abstention. We use structured-output decoding on backends that support the schema.

\subsubsection*{System Prompt}

In every prompt and prompt example shown in this appendix and in Appendix~\ref{appx:prompts}, the token ``TGN'' is used as the in-prompt nickname for our \tgl{} network. The system prompt has three components. \emph{(i)} \textbf{Role description}: the model is told it is a proactive assistant that receives the rolling session history, the current observation, and a routing list of graph-identified topics most relevant to the current activity. \emph{(ii)} \textbf{Task formulation}: treat the first entry of the routing list as the routing anchor and use its surface text as the artifact the user is working with; quote the anchor text inline using single quotes (e.g.\ \texttt{'email\_filter.py'}), preserving any extension or domain marker verbatim; lead with what the user \emph{receives} (a summary, an extract, a draft, a comparison, a checklist, a function added to the anchor) and formulate the task around the supplied artifact. The task is constrained to a single sentence of at most $150$ characters ending in a single period, and excludes the internal labels ``TGN'', ``routing'', ``graph'', or ``scores''. \emph{(iii)} \textbf{Three hand-written few-shot examples}, each a JSON input bundling the session history, current observation, and routing list, paired with a JSON output. The examples cover a coding session with a quoted file anchor, a research session whose routing is intentionally empty (showing how the model uses session history when routing is empty), and a writing session with a markdown file anchor.

The \emph{Ours} prompt is shared across all downstream backbones. \emph{No trigger} and \emph{Random routes} use that prompt, with the former invoking the agent at every event and the latter receiving a sampled routing list. \emph{No \tgl{}} retains the output-style directives and uses language-agent self-gating; its prompt omits the routing field, routing-specific instructions, and three routing demonstrations. Appendix~\ref{appx:anatomy-full} specifies the random-list construction.

\subsubsection*{Decoding}

Local vLLM backbones use temperature $0$, and API backbones use the provider's default sampling settings. The output cap is $1{,}536$ tokens. JSON output is requested through a structured-output schema where supported and through format instructions otherwise. The parsing procedure retries with \texttt{json\_object} mode and then unconstrained generation, followed by regex extraction of the first JSON object. If parsing fails throughout, the raw response is sent to the benchmark judge. Each \emph{Ours} run uses the same prompt content across backbones.

\subsection{Instantiation: Mobile Agent (FingerTip)}
\label{appx:agent:fingertip}

The mobile implementation predicts one intent per FingerTip episode. It uses \texttt{Qwen2.5-VL-7B-Instruct} with $t$ episode screenshots ($t$ is the number of images) and graph-based retrieval candidates. The semantic head retrieves training-set intent strings, which pass through the same-app filter before entering the language agent's context. The prompt requests a Chinese intent imperative of $8$--$25$ characters.

\subsubsection*{Candidate Retrieval}

We encode each test episode with \tgl{} and read the current-episode node's $1024$-dimensional $L_2$-normalised semantic vector. Training vectors are precomputed for all episodes in the released training split. Cosine similarity ranks the training neighbours for each test episode. We retain the top six neighbours and remove duplicate target-intent strings before same-app filtering.

\subsubsection*{Same-App Coherence Filter}

The same-app filter retains candidates whose app matches the top-ranked candidate, yielding at most two candidates. Confidence is high when the highest cosine similarity is at least $0.78$ and two candidates survive, low when the highest cosine similarity is below $0$, and medium otherwise. The $0.78$ threshold is fixed before evaluation. The test-set confidence distribution is $56\%$ high, $44\%$ medium, and $0\%$ low.

\subsubsection*{Prompt and Decoding}

The user payload bundles the user profile, time, scenario, the last $20$ previous intents, the retrieval-confidence label, and the retrieved intent candidates (each carrying an app, an intent class, and an intent text). When $t \geq 1$ the first $t$ episode screenshots are loaded as JPEG bytes, base64-encoded, and prepended to the user message as image content blocks. The system prompt (in Chinese, since the target intent strings are Chinese) instructs the model to read the screenshots first, then the previous intents, then the candidates as a soft prior whose use depends on the retrieval-confidence label: high-confidence candidates may be copied as-is; medium-confidence candidates are advisory, with the predicted app grounded in the screenshots and history; low-confidence candidates are ignored entirely. The prompt requires a JSON output containing the predicted app, the predicted intent class, and a proposed-task string of $8$ to $25$ Chinese characters containing the predicted app name, plus session-theme and thoughts bookkeeping fields. Four hand-written few-shot demonstrations precede the test payload: three demonstrate how high-confidence candidates are combined with recent history (an alarm-setting episode anchored on 闹钟 (Clock), a video-watching episode anchored on 哔哩哔哩 (Bilibili), a food-delivery episode anchored on 美团 (Meituan)), and one demonstrates a low-confidence case in which the model synthesises its answer from the user's repeated activity in the session history (the override demonstration is anchored on 支付宝 (Alipay) drawn from the user's repeated morning routine). The override demonstration specifies the low-confidence branch of the trust-gating rule.

The local vLLM agent uses temperature $0$ and a $320$-token output cap. Parsing strips Markdown fences and extracts the first JSON object. The proposed-task field must contain $1$--$60$ characters and pass a wrapper filter. This filter rejects the prefixes 用户 / 时间： / 地点： / 意图： and the pattern ``用户 \ldots 意图是 \ldots''. A failed parse or filter check returns the top-ranked retrieved intent. At least $99\%$ of outputs parse successfully under this configuration.

%% file: src/D_additional_experiments.tex
\section{Additional Experiments}
\label{appx:additional}
\suppressfloats[t]

\subsection{Ablation Study: Full Numbers}
\label{appx:anatomy-full}

Table~\ref{tab:broadrandom-ablation} gives the values plotted in Figure~\ref{fig:anatomy}, including variation across three runs. The controls address three parts of the pipeline: \emph{No \tgl{}} uses language-agent self-gating with the style-only prompt, \emph{No trigger} retains learned routing and invokes the agent at every event, and \emph{Random routes} retains the trigger and replaces the routed context. Each loss column reports how many F1 percentage points the control falls below \emph{Ours}. Figure~\ref{fig:anatomy} plots absolute F1 and annotates the differences from \emph{Ours}.

\paragraph{Random routing.} We sample an equal-length replacement list uniformly from the distinct entity labels in the training vocabulary. The pool includes files, applications, search queries, web addresses, and category nodes. Every sampled entity is supplied with relevance score $0.5$. Sampling is seeded by event, so repeated runs use the same replacement list. This control measures the effect of the supplied entity content under the deployed trigger and prompt.

\input{src/tables/broadrandom_ablation}

\subsection{Trigger-Architecture Baselines}
\label{appx:trigger-baselines}

Table~\ref{tab:trigger-arch} compares eight alternative triggers with \tgl{}. The classifier rows use event labels derived from the released training set; the generation-based SFT row uses released task conversations. \tgl{} also receives entity-routing supervision. Each trigger is evaluated with the same routing list, downstream prompt, and language-agent backbones. The inputs and optimisation settings below define the architecture comparison.

\paragraph{Rule.} A dictionary lookup that maps each event's verb to its empirical fire rate observed in the training event-trigger labels. At inference the trigger probability is the lookup value; the trigger threshold sweeps that value.

\paragraph{Tabular (LR, HGB).} Logistic regression (LR) and histogram gradient boosting (HGB) (\texttt{sklearn} defaults) fitted on a $65$-dimensional hand-crafted feature vector per event: verb identity (one-hot), foreground-application identity (one-hot), per-entity-type counts (files, applications, search queries, web addresses, and category nodes in the current event), and time-since-previous-event in the session.

\paragraph{Textual (BGE-frozen + MLP, BGE-FT + MLP).} A two-layer multilayer perceptron (MLP) head sits on the \texttt{[CLS]} representation of the last five session events, joined by \texttt{[SEP]} and encoded by a \texttt{BAAI/bge-base-en-v1.5} encoder. The frozen variant freezes the encoder and trains only the MLP head ($0.2$\,M parameters), isolating the encoder's pre-trained representation; the fine-tuned (FT) variant updates the encoder end-to-end ($109$\,M trainable). Both use the same trigger labels and positive-class weighting as \tgl{}.

\paragraph{LLM-as-trigger (Qwen3-0.6B, Qwen3-8B).} A chat-tuned LLM is asked one binary yes/no question per event with the rolling session history and current observation as user payload, and the trigger probability is read off the next-token logit as $P(\mathrm{yes})/(P(\mathrm{yes})+P(\mathrm{no}))$ summed over case-variant token IDs ($\{$yes, Yes, YES$\}$ vs.\ $\{$no, No, NO$\}$). Qwen3 chat models default to reasoning mode and emit a \texttt{<think>} token first (probability ${\approx}1$ on every event without bypass); we append the empty thinking block \texttt{<think>$\backslash$n$\backslash$n</think>$\backslash$n$\backslash$n} after the chat-template prefix to expose the answer logit. Low-rank adaptation (LoRA) uses adapter rank $16$, adapter scaling parameter $\alpha=32$, dropout $0$, and adapters on the attention query, key, value, and output projections, $2$ epochs at lr $10^{-4}$ with warmup ratio $0.03$ in BF16 at effective batch $32$. LLM-trigger training uses a $1{:}1$ balance of fire and no-fire events. Loss is cross-entropy on the assistant's first token only. At inference we use the merged-LoRA HuggingFace transformers eager mode (\texttt{peft.merge\_and\_unload()}, batch $1$, \texttt{torch.cuda.synchronize()} before/after each forward) for the latency measurement; merging the adapter recovers base-model latency to within $\pm 3$\,ms.

\paragraph{ProactiveAgent full-SFT replication (Proactive-Qwen3-0.6B).} A full-parameter supervised fine-tuning (SFT) replication of the ProactiveAgent release recipe at the $0.6$\,B scale: \texttt{Qwen/Qwen3-0.6B} ($752$\,M parameters, every weight trainable, embeddings included), multi-turn conversations from the paper's released \texttt{agent\_trainset.json}, AdamW at lr $10^{-5}$ with cosine schedule and warmup ratio $0.1$, effective batch $32$ (per-device $1 \times$ grad-accum $32$, matching the paper's $8$ GPU $\times$ per-device $1 \times$ grad-accum $4$), $10$ epochs, sequence length $4{,}096$, BF16, max grad norm $1.0$, assistant-tokens-only cross-entropy loss. Wall-clock $520$\,min on a single NVIDIA RTX A6000 ($48$\,GB). At inference the per-event protocol matches the released-checkpoint protocol of \citet{lu2025proactiveagent}: the system prompt is the paper's Role/Task/Format/Rules block, the per-event user payload is the rolling session history plus the current observation as a JSON \texttt{\{Instructions, Observation\}} object, and the trigger is derived from the generated \texttt{proposed\_task} field (non-empty $\Rightarrow$ prob $1.0$, empty $\Rightarrow$ $0.0$), averaged over $3$ inference seeds. Sampling follows the Qwen3 non-thinking defaults (sampling temperature $0.7$, nucleus probability $0.8$, top-$20$ token sampling, and a $512$-token cap; mean $139$ generated tokens per event). This row is the only LLM trigger in Table~\ref{tab:trigger-arch} that derives its decision from a full task-text generation; its Latency columns therefore measure decoding cost.

\begin{figure}[htbp]
    \centering
    \includegraphics[width=\textwidth]{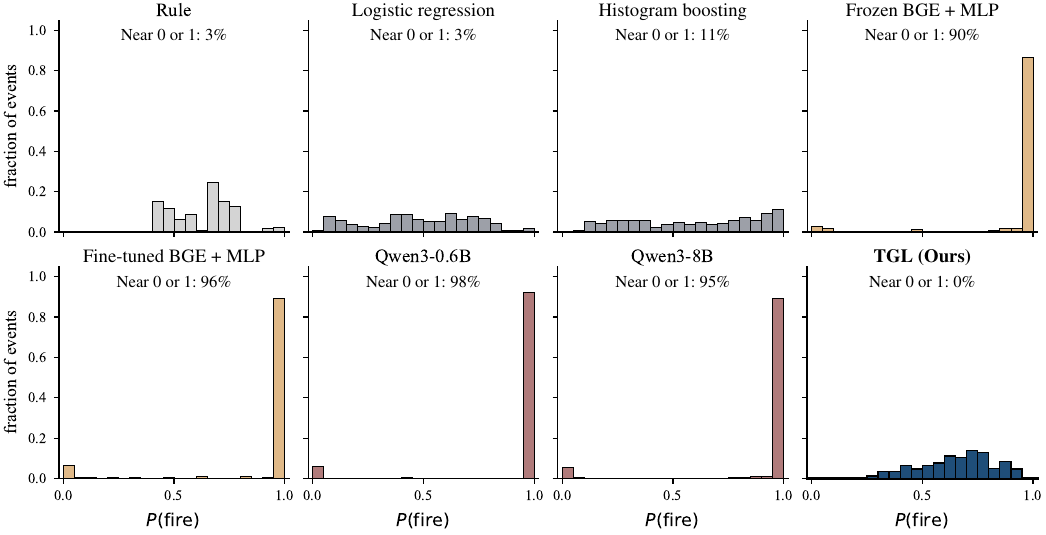}
    \caption{Distribution of trigger probabilities over desktop test events for eight models evaluated with one forward pass. Horizontal axes show the predicted probability of firing; vertical axes show the fraction of events in each bin of width $0.05$. Panel annotations give the percentage below $0.05$ or at least $0.95$. LR: logistic regression; HGB: histogram gradient boosting; MLP: multilayer perceptron. Fixed-threshold F1 is evaluated in Appendix~\ref{appx:rebuttal:threshold}.}
    \label{fig:score-distributions}
\end{figure}

\paragraph{Score distributions.} Figure~\ref{fig:score-distributions} shows the per-event trigger scores of the eight single-forward architectures. It describes their score scales alongside the ranking and calibration metrics in Table~\ref{tab:trigger-arch}.

\subsection{Drop-In \tgl{} Trigger on Fine-Tuned Downstream Agents}
\label{appx:dropin}

We also examine agents already fine-tuned for proactive assistance, testing whether graph-based filtering improves intervention selection while retaining their existing suggestions. We apply it offline to the two fine-tuned baselines from \citet{lu2025proactiveagent}, LLaMA-3.1-8B-Proactive and Qwen2-7B-Proactive: events whose trigger probability falls below the deployed threshold are forced to \texttt{task = null}; everything else (their published outputs, prompts, and decoding) is unchanged.

\begin{figure}[htbp]
    \centering
    \includegraphics[width=0.48\textwidth]{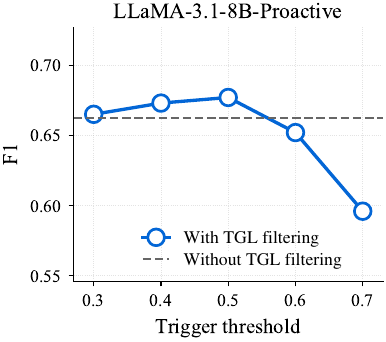}\hfill
    \includegraphics[width=0.48\textwidth]{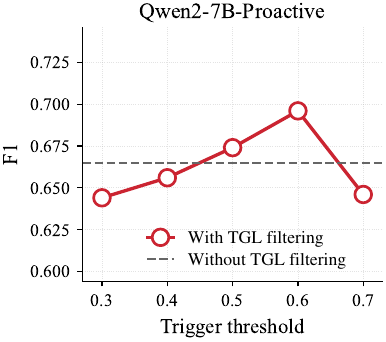}
    \caption{Filtering the outputs of two fine-tuned proactive agents. Coloured curves show F1 after applying the \tgl{} trigger at each threshold; grey dashed lines show the same fine-tuned agents without filtering. The deployed threshold is $0.5$. Outputs retained by the filter keep their original suggestions.}
    \label{fig:dropin-trigger}
\end{figure}

Figure~\ref{fig:dropin-trigger} plots F1 versus the applied trigger threshold, with a dashed line at the model's published F1 from Table~\ref{tab:main-results} ($66.25$ for LLaMA-3.1-8B-Proactive, $66.47$ for Qwen2-7B-Proactive). At our deployed threshold, both checkpoints exceed their published F1: LLaMA-3.1-8B-Proactive reaches $67.66$ ($+1.4$ F1) with $25\%$ fewer LLM calls, and Qwen2-7B-Proactive reaches $67.38$ ($+0.9$ F1) with $21\%$ fewer LLM calls. The sweep shows how the trade-off changes with the threshold: stronger filtering eventually removes useful interventions and lowers F1.

\subsection{Full Per-Backbone Main Results}
\label{appx:full-main-results}

Table~\ref{tab:main-results} gives every metric behind Figure~\ref{fig:main-results}: precision, recall, accuracy, false-alarm rate, and F1 for each of the $14$ backbones under the vanilla protocol and with \tgl{} triggering and routing, plus the two fine-tuned proactive baselines.

\input{src/tables/main_results}

\subsection{Per-Backbone Downstream-Agent Latency}
\label{appx:per-backbone-latency}

\input{src/tables/per_backbone_latency}

To quantify the generation cost avoided by skipping an event, Table~\ref{tab:per-backbone-latency} reports the wall-clock cost of one downstream-agent call for each of the $14$ backbones, observed end-to-end under our evaluation serving setup (open-weight on an A6000 vLLM; OpenAI / Anthropic / DeepSeek over their respective official APIs). They include API queueing and load, which is the per-event cost a deployed system pays. The fastest measured API call ($1503$\,ms for GPT-5.4-mini) is $\approx 135\times$ \tgl{}'s trigger forward.

%% file: src/tables/broadrandom_ablation.tex
\begin{table}[htbp]
\centering
\small
\setlength{\tabcolsep}{4pt}
\renewcommand{\arraystretch}{1.1}
\setlength{\abovecaptionskip}{0pt}
\caption{Pipeline ablations: F1 in percent (mean $\pm$ standard deviation over three runs). No \tgl{} uses language-agent self-gating and a style-only prompt; No trigger invokes the agent at every event; Random routes replaces the entity list with equally many sampled training entities. $\Delta = \mathrm{F1}_{\text{Ours}}-\mathrm{F1}_{\text{control}}$ is the F1 loss in percentage points.}
\label{tab:broadrandom-ablation}
\begin{tabular}{l rr rr rr r}
\toprule
                       & \multicolumn{2}{c}{No \tgl{}} & \multicolumn{2}{c}{No trigger} & \multicolumn{2}{c}{Random routes} & \\
\cmidrule(lr){2-3} \cmidrule(lr){4-5} \cmidrule(lr){6-7}
Agent                  & F1 & Loss & F1 & Loss & F1 & Loss & \textbf{Ours} \\
\midrule
\multicolumn{8}{l}{\emph{Local-deployable open-weight}} \\
Qwen2-7B-Instr.        & $44.54{\pm}3.11$ & $+26.1$ & $67.95{\pm}2.59$ & $+2.7$  & $56.12{\pm}3.52$ & $+14.6$ & $\mathbf{70.68{\pm}2.24}$ \\
LLaMA-3.1-8B-Instr.    & $49.63{\pm}2.76$ & $+22.4$ & $68.46{\pm}2.16$ & $+3.6$  & $51.38{\pm}6.09$ & $+20.7$ & $\mathbf{72.07{\pm}3.68}$ \\
Gemma-3-12B-it         & $64.82{\pm}1.98$ & $+12.5$ & $74.05{\pm}1.22$ & $+3.2$  & $70.40{\pm}1.67$ & $+6.9$  & $\mathbf{77.30{\pm}1.79}$ \\
Qwen3-4B               & $55.45{\pm}0.45$ & $+11.3$ & $62.51{\pm}2.13$ & $+4.3$  & $60.34{\pm}2.34$ & $+6.4$  & $\mathbf{66.77{\pm}3.04}$ \\
Qwen3-8B               & $55.32{\pm}1.90$ & $+16.8$ & $68.85{\pm}0.90$ & $+3.3$  & $53.99{\pm}0.56$ & $+18.1$ & $\mathbf{72.14{\pm}0.49}$ \\
\midrule
\multicolumn{8}{l}{\emph{Cloud/API}} \\
DeepSeek-V4-Flash      & $67.15{\pm}1.98$ & $+9.5$  & $73.66{\pm}3.72$ & $+2.9$  & $64.94{\pm}1.46$ & $+11.7$ & $\mathbf{76.60{\pm}3.33}$ \\
DeepSeek-V4-Pro        & $67.50{\pm}4.11$ & $+6.2$  & $71.05{\pm}4.01$ & $+2.7$  & $71.35{\pm}3.81$ & $+2.4$  & $\mathbf{73.70{\pm}4.81}$ \\
GPT-4o-mini            & $62.80{\pm}2.11$ & $+6.6$  & $65.96{\pm}0.89$ & $+3.4$  & $63.50{\pm}1.22$ & $+5.9$  & $\mathbf{69.38{\pm}1.33}$ \\
GPT-4o                 & $58.17{\pm}1.67$ & $+15.1$ & $70.65{\pm}1.99$ & $+2.6$  & $63.50{\pm}1.49$ & $+9.7$  & $\mathbf{73.23{\pm}2.64}$ \\
GPT-5.4-mini           & $67.14{\pm}1.90$ & $+10.4$ & $75.24{\pm}3.05$ & $+2.3$  & $69.35{\pm}2.73$ & $+8.2$  & $\mathbf{77.55{\pm}2.40}$ \\
GPT-5.4                & $73.78{\pm}0.62$ & $+2.8$  & $74.36{\pm}2.66$ & $+2.2$  & $69.53{\pm}0.51$ & $+7.0$  & $\mathbf{76.57{\pm}3.94}$ \\
Claude-Haiku-4.5       & $53.38{\pm}4.35$ & $+20.0$ & $70.47{\pm}0.88$ & $+2.9$  & $71.41{\pm}1.94$ & $+2.0$  & $\mathbf{73.40{\pm}0.96}$ \\
Claude-Sonnet-4.6      & $64.56{\pm}2.41$ & $+11.6$ & $70.91{\pm}4.35$ & $+5.3$  & $75.14{\pm}3.08$ & $+1.1$  & $\mathbf{76.20{\pm}3.35}$ \\
Claude-Opus-4.7        & $73.40{\pm}2.09$ & $+6.5$  & $75.86{\pm}1.52$ & $+4.0$  & $74.22{\pm}1.44$ & $+5.6$  & $\mathbf{79.86{\pm}1.57}$ \\
\midrule
mean                   & $61.26$              & $+12.7$ & $70.71$              & $+3.2$  & $65.37$              & $+8.6$  & $\mathbf{73.96}$ \\
\bottomrule
\end{tabular}
\end{table}

%% file: src/tables/main_results.tex
\begin{table}[htbp]
\centering
\small
\setlength{\tabcolsep}{6pt}
\setlength{\abovecaptionskip}{0pt}
\caption{Full desktop results in percent. Each backbone is evaluated with the benchmark's vanilla prompt and with \tgl{} triggering and routing (indented row). Bold marks the better value in each pair. Fine-tuned baselines and the first four vanilla rows are published results from \citet{lu2025proactiveagent}; the remaining vanilla rows are our re-runs with the same reward-model judge. Our runs average three evaluations (Appendix~\ref{appx:impl:seeds}). False alarm is the fraction of generated suggestions rejected by the judge.}
\label{tab:main-results}
\begin{tabular}{l rrrrr}
\toprule
Model & Precision $\uparrow$ & Recall $\uparrow$ & Accuracy $\uparrow$ & False alarm $\downarrow$ & F1 $\uparrow$ \\
\midrule
\multicolumn{6}{l}{\emph{Fine-tuned proactive baselines}} \\
LLaMA-3.1-8B-Proact.   & 49.76 &           99.06  & 52.86 & 50.24 & 66.25 \\
Qwen2-7B-Proact.       & 49.78 &          100.00  & 50.66 & 50.22 & 66.47 \\
\midrule
\multicolumn{6}{l}{\emph{Backbones with published vanilla results}} \\
Qwen2-7B-Instr.        &           44.00 &  \textbf{98.02} &           43.61 &           56.00 &           60.74 \\
\quad With \tgl{}          & \textbf{55.56} &           97.22 & \textbf{62.66} & \textbf{44.44} & \textbf{70.68} \\
LLaMA-3.1-8B-Instr.    &           38.16 &  \textbf{98.86} &           39.06 &           61.84 &           55.06 \\
\quad With \tgl{}          & \textbf{57.32} &           97.29 & \textbf{64.09} & \textbf{42.68} & \textbf{72.07} \\
GPT-4o                 &           48.15 &  \textbf{98.11} &           49.78 &           51.85 &           64.60 \\
\quad With \tgl{}          & \textbf{58.73} &           97.36 & \textbf{65.24} & \textbf{41.27} & \textbf{73.23} \\
GPT-4o-mini            &           35.28 & \textbf{100.00} &           36.12 &           64.73 &           52.15 \\
\quad With \tgl{}          & \textbf{53.97} &           97.14 & \textbf{61.37} & \textbf{46.03} & \textbf{69.38} \\
\midrule
\multicolumn{6}{l}{\emph{Backbones with re-run vanilla results}} \\
Gemma-3-12B-it         &           42.67 &           91.05 &           46.07 &           57.33 &           58.04 \\
\quad With \tgl{}          & \textbf{64.02} & \textbf{97.58} & \textbf{69.53} & \textbf{35.98} & \textbf{77.30} \\
Qwen3-4B               &           23.89 & \textbf{100.00} &           25.75 &           76.11 &           38.54 \\
\quad With \tgl{}          & \textbf{50.97} &           96.97 & \textbf{58.94} & \textbf{49.03} & \textbf{66.77} \\
Qwen3-8B               &           15.07 &  \textbf{99.07} &           19.17 &           84.93 &           26.14 \\
\quad With \tgl{}          & \textbf{57.32} &           97.31 & \textbf{64.09} & \textbf{42.68} & \textbf{72.14} \\
GPT-5.4-mini           &           52.97 &           94.74 &           63.23 &           47.03 &           67.94 \\
\quad With \tgl{}          & \textbf{64.37} & \textbf{97.59} & \textbf{69.81} & \textbf{35.63} & \textbf{77.55} \\
GPT-5.4                &           58.88 &  \textbf{97.72} &           64.38 &           41.12 &           73.47 \\
\quad With \tgl{}          & \textbf{63.14} &           97.54 & \textbf{68.81} & \textbf{36.86} & \textbf{76.57} \\
Claude-Haiku-4.5       &           45.03 &  \textbf{99.31} &           50.93 &           54.97 &           61.89 \\
\quad With \tgl{}          & \textbf{58.91} &           97.38 & \textbf{65.38} & \textbf{41.09} & \textbf{73.40} \\
Claude-Sonnet-4.6      &           49.87 &           71.23 &           58.85 &           50.13 &           58.64 \\
\quad With \tgl{}          & \textbf{62.61} & \textbf{97.52} & \textbf{68.38} & \textbf{37.39} & \textbf{76.20} \\
Claude-Opus-4.7        &           65.71 &           85.86 & \textbf{72.53} &           34.29 &           74.44 \\
\quad With \tgl{}          & \textbf{67.55} & \textbf{97.70} &           72.39 & \textbf{32.45} & \textbf{79.86} \\
DeepSeek-V4-Flash      &           51.25 &           83.73 &           57.94 &           48.75 &           63.44 \\
\quad With \tgl{}          & \textbf{63.14} & \textbf{97.54} & \textbf{68.81} & \textbf{36.86} & \textbf{76.60} \\
DeepSeek-V4-Pro        &           43.26 &           52.37 &           58.94 &           56.74 &           47.12 \\
\quad With \tgl{}          & \textbf{59.44} & \textbf{97.38} & \textbf{65.81} & \textbf{40.56} & \textbf{73.70} \\
\bottomrule
\end{tabular}
\end{table}

%% file: src/tables/per_backbone_latency.tex
\begin{table}[htbp]
\centering
\setlength{\tabcolsep}{10pt}
\setlength{\abovecaptionskip}{0pt}
\caption{Wall-clock latency of one downstream generation call and its ratio to the $11.13$\,ms \tgl{} controller latency. Measurements include request handling and API queueing. \emph{Serving:} open-weight backbones via A6000 vLLM; OpenAI / Anthropic / DeepSeek backbones via the respective vendors' official APIs.}
\label{tab:per-backbone-latency}
\begin{tabular}{l rr}
\toprule
Backbone               & Latency (ms/call) & Ratio to controller \\
\midrule
\multicolumn{3}{l}{\emph{Open-weight (A6000 vLLM)}} \\
Qwen2-7B-Instr.        & $2412$  & $217\times$ \\
LLaMA-3.1-8B-Instr.    & $2638$  & $237\times$ \\
Gemma-3-12B-it         & $4461$  & $401\times$ \\
Qwen3-4B               & $1800$  & $162\times$ \\
Qwen3-8B               & $2634$  & $237\times$ \\
\midrule
\multicolumn{3}{l}{\emph{Closed/API}} \\
DeepSeek-V4-Flash      & $5705$  & $513\times$ \\
DeepSeek-V4-Pro        & $21839$ & $1962\times$ \\
GPT-4o-mini            & $2387$  & $215\times$ \\
GPT-4o                 & $1688$  & $152\times$ \\
GPT-5.4-mini           & $1503$  & $135\times$ \\
GPT-5.4                & $2431$  & $218\times$ \\
Claude-Haiku-4.5       & $4366$  & $392\times$ \\
Claude-Sonnet-4.6      & $5714$  & $513\times$ \\
Claude-Opus-4.7        & $5676$  & $510\times$ \\
\bottomrule
\end{tabular}
\end{table}

%% file: src/M_rebuttal_analyses.tex
\section{Additional Robustness and Calibration Analyses}
\label{appx:rebuttal}

We evaluate backward temporal edges, fixed-threshold performance, graph-construction choices, and events with an empty routing list. Retraining follows Appendix~\ref{appx:impl}; downstream evaluations use five local open-weight backbones with three runs each. For the graph-structure analyses, trigger AUC uses the must-fire target from Section~\ref{sec:results:trigger-arch}, with $0.738$ for the deployed model. The must-fire label marks events requiring a non-null task in the released test annotations. Routing-anchor agreement is the fraction of events with extracted entities on which a variant selects the same highest-scoring entity as the deployed model; retraining the original recipe gives agreement of $0.90$.

\subsection{Backward-Temporal-Edge Ablation}
\label{appx:rebuttal:causal}

The training graph includes forward and backward temporal edges. The deployed inference procedure removes backward temporal edges and masks future-node input representations (Appendix~\ref{appx:causal}). We retrain with both edge configurations to examine how backward training edges affect triggering and routing under the deployed inference procedure. A separate variant adds backward edges at inference.

\paragraph{Setup.} We retrain in two arms: (a) \emph{both-direction training}, using earlier-to-later and later-to-earlier event edges; (b) \emph{forward-only training}, using earlier-to-later event edges. We use five training seeds per condition and evaluate both with the deployed inference procedure.

\paragraph{Trigger AUC.}
Table~\ref{tab:causal-auc} reports mean trigger AUC of $0.733$ for both training arms. The paired per-seed difference is $0.000\pm0.011$. The deployed checkpoint's AUC of $0.738$ lies within the retraining range of $0.722$--$0.744$.

\begin{table}[htbp]
\centering
\setlength{\tabcolsep}{10pt}
\setlength{\abovecaptionskip}{0pt}
\caption{Effect of temporal-edge direction during training. Trigger AUC uses the released must-fire labels. Retrained rows report mean $\pm$ standard deviation over five seeds; both use the deployed inference procedure.}
\label{tab:causal-auc}
\begin{tabular}{lc}
\toprule
Training configuration & Must-fire AUC $\uparrow$ \\
\midrule
Deployed model (Table~\ref{tab:trigger-arch}) & 0.738 \\
Both temporal directions & 0.733 $\pm$ 0.009 \\
Forward temporal edges only & 0.733 $\pm$ 0.008 \\
\midrule
Paired difference (forward-only $-$ both) & 0.000 $\pm$ 0.011 \\
\bottomrule
\end{tabular}
\end{table}

\paragraph{Routing-anchor agreement.}
The two training arms select the same top-ranked routing entity on $96\%$ of anchored events, measured with paired seeds. Retraining the original recipe yields $90\%$ reference agreement with the deployed model.

\paragraph{End-to-end F1.}
Each arm contributes one retrained checkpoint to the full pipeline, evaluated on five open-weight backbones with three downstream runs each. Table~\ref{tab:causal-f1} reports differences between the run means: removing backward edges changes mean F1 by $-0.5$ points. The two arms reach mean F1 of $71.9$ and $71.5$, compared with $71.8$ for the deployed checkpoint on these five backbones.

\begin{table}[htbp]
\centering
\setlength{\tabcolsep}{10pt}
\setlength{\abovecaptionskip}{0pt}
\caption{Effect of removing backward training edges on downstream F1. Each condition uses one retrained model with three runs per backbone. $\Delta$ is forward-only training minus both-direction training, in F1 percentage points.}
\label{tab:causal-f1}
\begin{tabular}{lc}
\toprule
Downstream backbone & $\Delta$ F1 (percentage points) \\
\midrule
Qwen2-7B-Instr & $-0.2$ \\
Qwen3-4B & $+1.3$ \\
Qwen3-8B & $+2.3$ \\
LLaMA-3.1-8B-Instr & $-4.1$ \\
Gemma-3-12B-it & $-1.7$ \\
\midrule
\textbf{mean} & $\mathbf{-0.5}$ \\
\bottomrule
\end{tabular}
\end{table}

\paragraph{Inference-edge variant.}
Adding backward temporal edges at inference changes the deployed checkpoint's trigger AUC by less than $0.005$, preserves the routing selections on every test event, and changes end-to-end F1 by $+0.03\pm0.3$ across five backbones and three runs.

\subsection{Calibration and Fixed-Threshold Reliability}
\label{appx:rebuttal:threshold}

\paragraph{Probability calibration.}
We evaluate whether the trigger's numerical probabilities match the frequency with which assistance is needed. For this analysis, the released reward-model judge evaluates an empty suggestion at every event: rejection of that suggestion gives label $1$ (fire), and acceptance gives label $0$ (skip). We fit the calibrator on judge-labelled training events and evaluate on judge-labelled test events, with positive rates of $55.25\%$ and approximately $34\%$, respectively. These rates use all judged events in the corresponding split as their denominator. The original trigger-training labels come from proposed tasks and have a positive rate of $88\%$ (Appendix~\ref{appx:graph:desktop}).

Temperature scaling adjusts the trigger's probability scale after training. Given its logit $z$, the calibrated probability is $p=\sigma(z/T)$, with fitted $T\approx3.27$. The positive temperature preserves score ordering and decisions at $\tau=0.5$. The Brier score is the mean squared error between probability and binary label. Expected calibration error (ECE) averages the absolute gap between predicted probability and observed positive rate over ten equally spaced probability bins, weighted by the number of events in each bin. Table~\ref{tab:recalibration} uses the judge-labelled fire/skip target for all three metrics: scaling reduces Brier from $0.310$ to $0.259$ and ECE from $0.336$ to $0.227$, while AUC remains $0.703$.

\begin{table}[htbp]
\centering
\small
\setlength{\tabcolsep}{6pt}
\setlength{\abovecaptionskip}{0pt}
\caption{Post-training calibration of trigger probabilities. Temperature scaling uses $p=\sigma(z/T)$, with positive $T$ fitted on judge-labelled training events. All metrics use the same judge-labelled fire/skip target on the test split: a positive label means the judge rejects an empty suggestion. The final column reports the percentage of test events whose decisions change after calibration at $\tau=0.5$.}
\label{tab:recalibration}
\begin{tabular}{lcccc}
\toprule
Probability mapping & AUC $\uparrow$ & Brier $\downarrow$ & ECE $\downarrow$ & \shortstack{Changed\\decisions} \\
\midrule
Uncalibrated & 0.703 & 0.310 & 0.336 & -- \\
Temperature-scaled & 0.703 & 0.259 & 0.227 & $0\%$ \\
\bottomrule
\end{tabular}
\end{table}

\paragraph{Threshold sensitivity.}
We vary the threshold around the deployed $\tau=0.5$ to examine how sensitive suggestion quality is to the operating point. Across five open-weight backbones with three runs each, mean F1 ranges from $70.4$ to $73.1$ over $\tau\in[0.4,0.6]$, with $72.1$ at the deployed threshold. The wider sweep $\tau\in[0.3,0.7]$ yields $67.8$--$73.5$. Post-hoc per-run threshold selection on the test set improves mean F1 by at most $1.5$ points over the deployed threshold.

\paragraph{Cross-backbone threshold transfer.}
To assess threshold reuse when changing the language agent, we choose the threshold maximising mean F1 on four open-weight backbones and apply it to the held-out fifth, rotating through all five. This analysis uses the same benchmark events across backbones. The transferred threshold incurs a mean loss of $0.4$ F1 relative to each held-out backbone's own tuned threshold (Table~\ref{tab:threshold-transfer}).

\begin{table}[htbp]
\centering
\setlength{\tabcolsep}{10pt}
\setlength{\abovecaptionskip}{0pt}
\caption{Threshold transfer across five backbones on the same test events. A threshold is selected using four backbones and applied to the fifth. $\Delta$ F1 is transferred-threshold F1 minus the held-out backbone's individually tuned F1, in percentage points; each mean uses three downstream runs.}
\label{tab:threshold-transfer}
\begin{tabular}{lc}
\toprule
Held-out backbone & $\Delta$ F1 (percentage points) \\
\midrule
Qwen2-7B & $-0.2$ \\
Qwen3-4B & $-0.5$ \\
Qwen3-8B & $-0.1$ \\
LLaMA-3.1-8B & $-1.1$ \\
Gemma-3-12B & $-0.2$ \\
\midrule
\textbf{mean} & $\mathbf{-0.4}$ \\
\bottomrule
\end{tabular}
\end{table}

\subsection{Construction-Rule Sensitivity}
\label{appx:rebuttal:construction}

Graph construction determines both the routing candidates and message-passing context. We vary entity categories and feature channels to test sensitivity to these choices; random entity deletion tests sensitivity to missing context. Category nodes encode file extension, file category, website domain, or query language (Appendix~\ref{appx:graph:desktop}). Each node-removal variant removes the named nodes and their incident edges; each feature variant disables the named embedding. Random deletion removes $25\%$ of entities. We retrain the controller for each variant. The variants use three training seeds and the deployed inference procedure; the reference and entity-deletion rows reuse the existing retraining runs. Table~\ref{tab:construction-sensitivity} reports trigger AUC for decision quality and anchor agreement for changes in entity selection.

\begin{table}[htbp]
\centering
\setlength{\tabcolsep}{9pt}
\setlength{\abovecaptionskip}{0pt}
\caption{Sensitivity to graph construction. AUC measures trigger ranking against must-fire labels (mean $\pm$ standard deviation across training seeds). Routing agreement is the fraction of events with entities where the highest-ranked entity matches the deployed model. All rows use the deployed inference procedure.}
\label{tab:construction-sensitivity}
\begin{tabular}{p{0.53\linewidth}cc}
\toprule
Construction change & Must-fire AUC $\uparrow$ & \shortstack{Routing\\agreement} \\
\midrule
\textbf{Original construction} & 0.733 $\pm$ 0.009 & 0.90 \\
Remove all category nodes & 0.717 $\pm$ 0.012 & 0.78 \\
Remove search-query and language-category nodes & 0.744 $\pm$ 0.008 & 0.86 \\
Remove web-address and domain nodes & 0.735 $\pm$ 0.005 & 0.89 \\
Disable node-type embedding & 0.725 $\pm$ 0.004 & 0.83 \\
Disable time-gap embedding & 0.715 $\pm$ 0.016 & 0.89 \\
Randomly remove 25\% of entities & 0.725 $\pm$ 0.019 & 0.71 \\
\bottomrule
\end{tabular}
\end{table}

\paragraph{Trigger prediction.} Mean AUC ranges from $0.715$ to $0.744$ across the construction variants, compared with $0.733\pm0.009$ for the reference. Disabling the time-gap embedding gives $0.715\pm0.016$, removing category nodes gives $0.717\pm0.012$, and random deletion of $25\%$ of entities gives $0.725\pm0.019$.

\paragraph{Routing selections.} Anchor agreement ranges from $0.83$ to $0.89$ when individual entity classes or feature channels are removed. Removing category nodes gives agreement of $0.78$, and random entity deletion gives $0.71$. These measurements describe the routing selections under each change to the graph's candidate entities and features.

\subsection{Events with No Extracted Entities}
\label{appx:rebuttal:zero-entity}

The controller processes every observation, including those with no extracted entities to route. We therefore examine trigger ranking and suggestion quality separately for events with and without extracted entities. At $34\%$ of test events, the current observation yields an empty entity set and the routing list is empty. For example, \emph{``The user scrolls and clicks on a research report page about biodegradable cups.''} contributes its text and temporal position to the session graph. Table~\ref{tab:zero-entity} reports both groups for five open-weight backbones with three downstream runs each.

\begin{table}[htbp]
\centering
\small
\setlength{\tabcolsep}{4pt}
\setlength{\abovecaptionskip}{0pt}
\caption{Performance by entity availability in the current observation. Event share uses all test events; must-fire rate is the positive-label fraction within each group. Predicted fire rate uses $\tau=0.5$. End-to-end F1 reports mean $\pm$ standard deviation across five backbones with three runs each; full event text and history remain available to both groups.}
\label{tab:zero-entity}
\begin{tabular}{lccccc}
\toprule
Current observation & \shortstack{share of\\events} & \shortstack{must-fire\\rate} & \shortstack{Trigger AUC\\(must-fire)} & \shortstack{Predicted fire\\rate at $\tau{=}0.5$} & \shortstack{end-to-end\\F1} \\
\midrule
No entities extracted & 34\% & 0.19 & \textbf{0.789} & 59\% & \textbf{75.0 $\pm$ 6.8} \\
Entities extracted & 66\% & 0.26 & 0.723 & 92\% & 71.0 $\pm$ 4.4 \\
\bottomrule
\end{tabular}
\end{table}

\paragraph{Triggering from event text and history.}
The event readout aggregates its text, timing, and neighbouring session context. Trigger AUC is $0.789$ on events with no extracted entities and $0.723$ on events with extracted entities, with fire rates of $59\%$ and $92\%$. Within one session, the report-scrolling event above fires with $p=0.73$ and carries a must-fire label. A later housekeeping event, \emph{``The user types `cmdw' to close a tab and continues with scrolls and clicks.''}, receives $p=0.27$ and is skipped.

\paragraph{Generating suggestions from session history.}
For an empty routing list, the downstream prompt uses the session history to formulate the task (Appendix~\ref{appx:prompts}). Mean end-to-end F1 is $75.0$ on events with no extracted entities and $71.0$ on events with extracted entities.

%% file: src/E_implementation_details.tex
\section{Implementation Details}
\label{appx:impl}

This appendix lists the architectural and training details deferred from Section~\ref{sec:method} and Section~\ref{sec:experiment}.

\subsection{Evaluation Backbones}
\label{appx:backbones}

Fourteen instruction-following backbones serve as the downstream language agent in the panel evaluations of Table~\ref{tab:main-results} and the ablations of Tables~\ref{tab:broadrandom-ablation} and \ref{tab:trigger-arch}:
\begin{itemize}
\item \emph{Open-weight (5)}, served via vLLM on a single A6000: Qwen2-7B-Instruct, LLaMA-3.1-8B-Instruct, Gemma-3-12B-it, Qwen3-4B, Qwen3-8B.
\item \emph{OpenAI (4)}, served via the OpenAI API: GPT-4o-mini, GPT-4o, GPT-5.4-mini, GPT-5.4.
\item \emph{Anthropic (3)}, served via the Anthropic API: Claude-Haiku-4.5, Claude-Sonnet-4.6, Claude-Opus-4.7.
\item \emph{DeepSeek (2)}, served via the DeepSeek API: DeepSeek-V4-Flash, DeepSeek-V4-Pro.
\end{itemize}
Two fine-tuned proactive baselines from \citet{lu2025proactiveagent} (LLaMA-3.1-8B-Proactive and Qwen2-7B-Proactive) appear for reference in Table~\ref{tab:main-results} and Figure~\ref{fig:dropin-trigger}; both are open-weight checkpoints with published reference results.

\paragraph{RM judge.}
All ProactiveAgent precision, recall, accuracy, false-alarm, and F1 numbers are computed with the released reward-model judge from \citet{lu2025proactiveagent}. The originating paper trains this RM on human-annotated event--prediction judgments and reports above-$90$ agreement with held-out human annotations in its reward-model assessment ($91.80$ F1 and $91.67$ accuracy).

\paragraph{Run aggregation.}\label{appx:impl:seeds}
Our downstream evaluation cells aggregate $n{=}3$ runs: provider-default sampling for API backbones and temperature $0$ for local vLLM backbones (run seeds $\{0,1,100\}$). Published legacy vanilla and fine-tuned reference values retain the originating paper's aggregation, as marked in Table~\ref{tab:main-results}. Each trigger-architecture metric in Table~\ref{tab:trigger-arch} (AUC, Brier, ECE, threshold standard deviation) is likewise averaged over $3$ inference runs (seeds $\{0, 1, 100\}$) of the same trained checkpoint; the trained trigger checkpoint is fixed across these inference runs.

\input{src/tables/trigger_resources}

\subsection{Node Featurisation}

Desktop nodes have three input features. \emph{(i)} \textbf{Text embedding:} a $768$-dimensional vector from the frozen \texttt{BAAI/bge-base-en-v1.5} encoder. Event nodes use the full event text, named entities use their surface labels, and category nodes use a serialised type--value label. For example, the Python-extension category is encoded as \texttt{file\_ext:py}. This literal string specifies the encoder input. \emph{(ii)} \textbf{Node-type embedding:} a learned $32$-dimensional vector identifying the node as an event, application, file, web address, search query, document artifact, file extension, file category, query-language category, or website domain. \emph{(iii)} \textbf{Time-gap embedding:} a $32$-dimensional multilayer perceptron (MLP) representation of $\log(1+s)$, where $s$ is the number of seconds since the preceding event. The first event and entity nodes use $s=0$.

The frozen text encoder runs once per minibatch. Concatenating the three features gives an $832$-dimensional vector, which a linear projection maps to the $256$-dimensional graph representation. Appendix~\ref{appx:graph:fingertip} specifies the mobile node vocabulary and Chinese text encoder.

\subsection{GNN Body}

Message passing runs for three rounds of relation-aware GATv2 attention \citep{brody2022attentivegraphattentionnetworks} (implemented as \texttt{torch\_geometric.nn.GATv2Conv}), each with: \emph{(i)} $4$ attention heads, head dimension $256/4 = 64$, concatenated; \emph{(ii)} a $32$-dimensional edge embedding (one entry per edge relation type) fed into each layer's attention; \emph{(iii)} explicit \texttt{self} edges with learned relation embeddings, supplied to GATv2Conv with \texttt{add\_self\_loops=False}; and \emph{(iv)} a residual + LayerNorm + GELU + dropout $0.2$ block around each layer, $x \leftarrow \text{LN}(x + \text{Drop}(\text{GELU}(\text{GATv2}(x, e, \text{edge\_emb}))))$.

\subsection{Combining Graph Layers}

Following \citet{xu2018jumping}, we keep the output of the input projection plus every GNN layer:
\begin{equation*}
h_{\text{jk}} = \text{concat}(\text{layer}_0, \text{layer}_1, \text{layer}_2, \text{layer}_3),
\end{equation*}
of dimension $256 \times 4 = 1024$. Both heads receive the node's input features and the context aggregated at each graph layer.

\subsection{Heads}

Two parallel MLPs, each \texttt{Linear($1024 \to 256$)} $\to$ \texttt{GELU} $\to$ \texttt{Dropout($0.2$)} $\to$ \texttt{Linear($256 \to 1$)}: the \textbf{trigger head} is applied only to event nodes and outputs the trigger logit; the \textbf{routing head} is applied only to entity nodes and outputs the per-entity relevance logit. Total trainable parameters: $1.16$\,M.

\subsection{Training}

Each batch produces one trigger logit per event and one routing logit per entity. We mask both outputs to nodes that have a label and use binary cross-entropy (BCE) losses computed from the logits:
\begin{align*}
\mathcal{L}_{\text{trig}} &= \text{BCE}(\text{logit}_{\text{trig}},\, y_{\text{trig}};\, w_{+} = 0.15), \\
\mathcal{L}_{\text{rout}} &= \text{BCE}(\text{logit}_{\text{rout}},\, y_{\text{rout}}), \\
\mathcal{L} &= 0.3\,\mathcal{L}_{\text{trig}} + 1.0\,\mathcal{L}_{\text{rout}}.
\end{align*}

\noindent
\textbf{Notes.} A label of $-1$ flags an unlabelled node and is excluded from the loss. The trigger positive-class weight $w_{+} = 0.15$ down-weights the dominant positive class in the training set. The relative loss weighting in Eq.~\eqref{eq:loss} is fixed at $\lambda_{\mathrm{trig}} = 0.3$ and $\lambda_{\mathrm{rout}} = 1.0$; the hidden width is $d = 256$ with $K = 4$ attention heads.

\textbf{Optimisation.} AdamW with learning rate $3 \times 10^{-4}$, weight decay $10^{-4}$, gradient clipping at global norm $1.0$. Each gradient step batches $32$ session graphs as disjoint graph components. We use the epoch-$2$ checkpoint with the fixed inference settings in Appendix~\ref{appx:inference-hparams}.

\textbf{Hardware and runtime.} All measurements run on a host with $8$ NVIDIA RTX A6000 GPUs ($48$\,GB each), $2 \times$ Intel Xeon Gold 6254 CPUs ($72$ threads at $3.10$\,GHz), and $251$\,GiB RAM. Training and per-event latency benchmarks use a single A6000. One epoch over the training set is $\approx 140$\,s wall-time, dominated by the BGE forward; training to $2$ epochs takes about $5$ minutes total.

\subsection{Inference on Event Histories}
\label{appx:causal}

Training encodes complete sessions with forward and backward temporal edges. For each event, we evaluate the session graph with backward temporal edges removed and future-node input representations masked. Trigger scores are read at $e_t$, and routing scores are read for entities connected to $e_t$. Appendix~\ref{appx:rebuttal:causal} evaluates the backward-edge channel under the deployed inference procedure.

\subsection{Inference Hyperparameters}
\label{appx:inference-hparams}

Trigger threshold $\tau_{\mathrm{trig}} = 0.5$ (the sigmoid decision boundary); routed-list cap $k = 20$. These hyperparameters are held fixed across all backbones and ablation cells.

\subsection{Latency Measurement Protocol}
\label{appx:latency-protocol}

The Latency columns of Table~\ref{tab:trigger-arch} are measured under two hardware regimes. \emph{Server} runs on a single NVIDIA A6000; the Rule, LR, and HGB rows run on the host's dual Intel Xeon Gold 6254 CPUs. \emph{Local} runs on an Apple M3 Pro laptop with $36$\,GB unified memory: Textual, LLM, and Graph rows execute under Metal Performance Shaders (MPS); Rule, LR, and HGB run on the M3 Pro CPU. Transformer inference uses auto padding under streaming-serve protocol (events arrive one at a time, the encoder is called per event), with the same dynamic-padding policy across transformer rows. Each cell is the mean of the last $3$ of $4$ timed forward passes; the first pass is discarded to absorb MPS compilation. All rows use eager inference at batch size one; the \tgl{} row uses the cached streaming protocol in Section~\ref{appx:cached-serving}, reusing recurring entity embeddings while encoding the event text and fresh entity strings per event. The Qwen3 LoRA rows on Local report the mean across the same three inference seeds, identical recipe to the server-side measurement. The Proactive-Qwen3-0.6B (full SFT) row is the one exception to single-forward measurement: deriving its binary trigger requires generating the full proposed-task text and then thresholding on whether the generation is non-empty, so each event triggers a full decoding pass (mean $139$ generated tokens; sampling temperature $0.7$, nucleus probability $0.8$, top-$20$ token sampling, and a $512$-token cap). Its latency cells therefore report full decoding cost: $3{,}927$\,ms on server and $11{,}966$\,ms on local, averaged across $3$ inference seeds.

\subsection{Cached Streaming Serving}
\label{appx:cached-serving}

The latency reported in Table~\ref{tab:trigger-arch} for \tgl{} ($11.13$\,ms server, $13.99$\,ms local) uses cached streaming serving under HuggingFace transformers + PyG inference: recurring entity embeddings are reused, while the event text and fresh entity strings are encoded per event. The measurement uses batch-$1$ streaming inference with no compilation, quantization, or cross-event batching.

The graph representation supports entity-level embedding caching during streaming inference. The BGE encoder is the dominant forward cost ($\sim\!7$--$8$ of the $11.13$\,ms), and most entity nodes a typical session touches are drawn from a small recurring vocabulary (applications, file extensions, file categories, query-language categories, website domains). Caching these embeddings reduces per-event BGE work to encoding only the event text plus any fresh entity strings (specific filenames, search queries, web addresses).

\begin{table}[htbp]
\centering
\small
\setlength{\tabcolsep}{6pt}
\renewcommand{\arraystretch}{1.05}
\setlength{\abovecaptionskip}{0pt}
\caption{Text-embedding computation during streaming inference. Known application and category labels can be encoded before streaming. New event text and unseen entity strings are encoded when encountered; recurring entity strings reuse their cached embeddings.}
\label{tab:caching-opportunity}
\begin{tabular}{p{0.32\linewidth}p{0.57\linewidth}}
\toprule
Node text & When the embedding is computed \\
\midrule
Event description & At each incoming event \\
File name & At first appearance; reused on recurrence \\
Web address & At first appearance; reused on recurrence \\
Search query & At first appearance; reused on recurrence \\
\midrule
Application name & Precomputed for known applications \\
File extension & Precomputed for the extension inventory \\
File category & Precomputed for the five file categories \\
Query-language category & Precomputed for the four script categories \\
Website domain & Precomputed for known deployment domains \\
\bottomrule
\end{tabular}
\end{table}

%% file: src/tables/trigger_resources.tex
\begin{table}[htbp]
\centering
\small
\setlength{\tabcolsep}{10pt}
\setlength{\abovecaptionskip}{0pt}
\caption{Resources used by the trigger models in Table~\ref{tab:trigger-arch}. Total parameters include frozen encoders; training time is wall-clock time. LR: logistic regression; HGB: histogram gradient boosting; MLP: multilayer perceptron; FT: fine-tuning; SFT: supervised fine-tuning. Tree models have no neural parameter count.}
\label{tab:trigger-resources}
\begin{tabular}{l rr r}
\toprule
Trigger                  & Total parameters       & Trainable      & Training time \\
\midrule
Rule                     & $0$                & $0$            & $<\!1$\,s \\
LR ($65$ features)       & $65$               & $65$           & $41$\,s \\
HGB ($65$ features)      & --                 & --             & $4$\,s \\
BGE-frozen + MLP         & $109.5$\,M         & $0.2$\,M       & ${\sim}5$\,s \\
BGE-FT + MLP             & $109.5$\,M         & $109.5$\,M     & $23$\,min \\
Qwen3-0.6B               & $757$\,M           & $4.59$\,M      & $19$\,min \\
Qwen3-8B                 & $8.21$\,B          & $15.34$\,M     & $86$\,min \\
Proactive-Qwen3-0.6B (full SFT) & $752$\,M & $752$\,M & $520$\,min \\
\textbf{\tgl{} (Ours)}   & $\mathbf{110.4}$\,\textbf{M} & $\mathbf{1.16}$\,\textbf{M} & $\mathbf{{\sim}5}$\,\textbf{min} \\
\bottomrule
\end{tabular}
\end{table}

%% file: src/F_prompts.tex
\section{System Prompts}
\label{appx:prompts}

This appendix reproduces the system prompts used by the downstream agent on each benchmark. Appendix~\ref{appx:agent} describes the input and output formats, few-shot settings, and decoding. The literal prompts retain the implementation nickname TGN for \tgl{}. In desktop inputs, \texttt{session\_history} contains recent event texts, \texttt{current\_observation} contains the current event, and \texttt{tgn\_routing\_nodes} contains the ranked entity list; each entry has its surface label (\texttt{node}) and relevance score (\texttt{score}). In mobile inputs, \texttt{tgn\_candidates} contains retrieved training intents, and \texttt{tgn\_confidence} is the high/medium/low retrieval-confidence label defined in Appendix~\ref{appx:agent:fingertip}. The output field \texttt{task} is the suggestion submitted for evaluation.

\subsection{Desktop ProactiveAgent}
\label{appx:prompts:desktop}

\subsubsection*{System prompt}

\begin{Verbatim}[fontsize=\scriptsize, breaklines=true, breakanywhere=true]
You are a proactive assistant. You are shown:
1. session_history: the last few user activities in this session (oldest first)
2. current_observation: what the user is doing right now
3. tgn_routing_nodes: graph-identified topics most relevant to the current activity

Your task: infer the user's current TASK THEME from the session history,
then generate ONE short, specific, actionable suggestion grounded in that theme.

Rules:
- Use session_history to identify what the user is working on (the session theme).
- If tgn_routing_nodes contains specific items (files, topics), anchor your task
  to one of them.
- The task must be a single sentence, concrete, max 150 characters.
- Reference the session theme explicitly (e.g., the programming language,
  document topic, tool).
- Do NOT give a generic suggestion like 'open the last file' or 'check recent
  activity'.
- Do NOT mention TGN, routing, graph, or scores.

TGN-grounded task formulation:
- The first entry of tgn_routing_nodes is the routing anchor: the entity the
  graph identified as the user's current focus. Build the task around delivering
  something concrete about that anchor; quote its surface text inline
  (e.g. 'email_filter.py') and preserve any file extension or domain marker
  verbatim.
- Lead with what the user RECEIVES, not what they should DO next. Because the
  routing anchor has already located the entity, do not mirror the user's last
  observed verb (open / search / browse) back as the task. Instead state the
  deliverable: a summary, an extract, a draft, a comparison, a checklist, a
  function added to the anchor.
- End the task with a single period. No question mark, no exclamation, no
  trailing offer or hedge: the routing score already encodes confidence, the
  surface text should match.

Output ONLY JSON: {"task":"string","session_theme":"string","purpose":"string",
"thoughts":"string","response":"string","intent":"string"}.
\end{Verbatim}

\subsubsection*{No-\tgl{} Prompt Variant (Style Only)}
\label{appx:prompts:no-tgn}

The \emph{No \tgl{}} ablation in Table~\ref{tab:broadrandom-ablation} replaces the production system prompt with the variant below. It retains the style directives (single sentence, $150$-character cap, lead-with-deliverable, do-not-invent-artifacts) and removes every reference to \texttt{tgn\_routing\_nodes}, the routing-anchor instructions, and all three few-shot demonstrations. The trigger is disabled; the agent self-gates by emitting \texttt{task = ""}.

\begin{Verbatim}[fontsize=\scriptsize, breaklines=true, breakanywhere=true]
You are a proactive assistant. You are shown:
1. session_history: the last few user activities in this session, oldest first.
2. current_observation: what the user is doing right now.

Your task: infer the user's current TASK THEME from the session history,
then generate ONE short, specific, actionable suggestion grounded in that theme.

Rules:
- Use session_history to identify what the user is working on.
- The task must be a single sentence, concrete, max 150 characters.
- Reference the session theme explicitly.
- Do NOT give a generic suggestion like "open the last file" or "check recent activity".
- Do NOT invent files, URLs, pages, or topics not present in the input.
- Lead with what the user RECEIVES, not what they should DO next.
- End non-empty tasks with a single period.
- If the current moment does not support a useful, specific suggestion, set task to "".

Output ONLY JSON: {"task":"string","session_theme":"string","purpose":"string",
"thoughts":"string","response":"string","intent":"string"}.
\end{Verbatim}

\subsection{FingerTip Mobile Proactive Agent}
\label{appx:prompts:fingertip}

The FingerTip prompt requests a Chinese intent string. Each candidate supplies an application, an intent class, and an intent description; the confidence label tells the agent how to combine candidates with screenshots and history.

\subsubsection*{System prompt}

\begin{Verbatim}[fontsize=\scriptsize, breaklines=true, breakanywhere=true]
你是一个安卓助手，根据用户画像、时间、场景、最近意图、（如果有）当前屏幕截图，
以及 TGN 时序图谱召回的候选意图，预测用户当前一句简短的中文意图。

输入字段说明：
- previous_intents: 该用户最近的意图（按时间升序）。是用户长期偏好的强信号。
- screenshot: 当前打开的 App 与界面，是**最强证据**（如果有）。
- tgn_candidates: 图谱召回的候选意图（按相关度排序）。**软先验，可信度由
  tgn_confidence 给出**：
  - high: 候选可信，强烈建议参考；可能直接 copy 一句作为 task。
  - medium: 候选有用但需谨慎；选 App 应跟截图/历史一致。
  - low: 候选不可信，**忽略 tgn_candidates**，只看截图与历史。
- 候选 App 跟截图/历史所示 App 不一致时，**永远以截图/历史为准**，可以独立生成意图。

输出严格 JSON：
{
  "session_theme": "20 字内总结用户最近在做什么 (例: '早晨重复设闹钟'、'晚上看哔哩哔哩')",
  "thoughts": "20 字内说明你怎么定下 task 的",
  "app": "用户当前使用的 App 中文名（要跟截图/历史/候选三者最强证据一致）",
  "intent_class": "用户行为大类（如 闹钟/计时、视频、商品搜索与购物、信息查询...）",
  "task": "一句 8-25 个汉字的中文祈使句，以动词或 App 名开头，包含 app 字段中的
          App 名，禁止 用户/时间：/地点：/意图： 前缀，禁止日期/星期/画像信息"
}

只输出 JSON，不要 markdown 包裹、不要前后说明。
\end{Verbatim}

%% file: src/G_case_study.tex
\section{Case Study}
\label{appx:case-study}

To inspect how routed context enters the generated task, we compare suggestions for the same observation. We pick three test events where all three pipelines (the benchmark's vanilla language agent, \emph{Ours}, \emph{Random routes}) fire and produce \emph{different} tasks. Each case shows the routing context the downstream agent received under \emph{Ours}, and the reward-model (RM) verdict for every variant. The downstream agent is Qwen2-7B-Instruct under the deployed configuration; \emph{Random routes} replaces the routing list (up to $20$ entries) with draws from the training entity pool (Section~\ref{sec:results:anatomy}).

\noindent\textbf{Case 1: coding session, romanised-Chinese query.} \tgl{} routes to the query-language category labelled English and the search query \emph{dianz youjian dizhi jiaoyan4 python}. The language category follows the Latin-script heuristic defined in Appendix~\ref{appx:graph:desktop}. Vanilla emits ``\emph{Translate the message from Chinese to English.}'' (rejected). \emph{Ours} quotes the query verbatim and ties it to the working file: ``\emph{Translate `dianz youjian dizhi jiaoyan4 python' into English for context in the `aw\_tools.py' file.}'' (accepted). \emph{Random routes} proposes a summary of the file from session history: ``\emph{Extract key points from the `aw\_tools.py' file and compile them into a concise summary.}'' (rejected). The accepted suggestion explicitly names both the routed query and the file from session history.

\noindent\textbf{Case 2: writing session, file and category context.} \tgl{}'s top-ranked routing entries for this event include the file \texttt{brandResearch.md} along with its writing-file category, Markdown-extension category, and the Visual Studio Code application. \emph{Ours} grounds on the file: ``\emph{Review and refine the structure of `brandResearch.md' for better readability.}'' (accepted). Vanilla proposes a research task: ``\emph{Search for key metrics and indicators used in measuring the sustainability of brands and provide them to the user for reference.}'' (rejected). \emph{Random routes} uses the file from session history and proposes an audience-engagement section: ``\emph{Add a section on audience engagement techniques to the `brandResearch.md' file.}'' (rejected).

\noindent\textbf{Case 3: application-only routing.} \tgl{} routes to \texttt{[VSCode]} only and defers the artifact to session history. \emph{Ours} produces a deliverable aligned with the user's note-taking activity: ``\emph{Extract and summarize Katla's product pricing details for your Markdown notes.}'' (accepted). Vanilla proposes a product comparison: ``\emph{Offer to compare the pricing of Katla's products with similar items from other eco-friendly brands, providing insights on positioning.}'' (rejected). \emph{Random routes} proposes a comparison table: ``\emph{Add a table comparing prices of Katla's eco-friendly products in your Markdown notes.}'' (rejected). The generated suggestion refers to the pricing information and Markdown notes present in the session context.

The three cases show how the downstream agent combines routed entities with session history. Each variant's quoted output and RM verdict are shown together, allowing the task formulation to be compared on the same observation.

%% file: src/H_discussion.tex
\section{Discussion}
\label{appx:discussion}

\paragraph{Graph structure and continuous control.} Events, recurring entities, and temporal connections provide the controller's representation. Shared message passing supplies an event-level trigger and entity-level routing scores in one forward pass. The two heads train jointly, and their outputs form a common interface for downstream agents. The routing-content ablation measures the interface's effect: the learned routing lists improve mean F1 by $8.6$ points over equal-length random lists under the same trigger and prompt.

\paragraph{Resource and latency implications.} The controller runs at $11.13$\,ms per event on an A6000 and $13.99$\,ms on an M3 Pro. Its runtime combines BGE text encoding with message passing over the session graph. Recurring entity embeddings are cached during streaming inference. The ${\sim}220$\,MiB BF16 resident footprint supports local activity processing, and triggered events receive downstream language-agent generation. Section~\ref{sec:efficiency} details monitoring and downstream generation costs.

\paragraph{Compatibility with proactive checkpoints.} The trigger also serves as a pre-filter for existing proactive language agents. Appendix~\ref{appx:dropin} applies it to the published outputs of LLaMA-3.1-8B-Proactive and Qwen2-7B-Proactive. At the deployed threshold, F1 increases by $1.4$ and $0.9$ points, respectively, and the corresponding call reductions are $25\%$ and $21\%$. The retained events use the original checkpoints' suggestions.